\documentclass[runningheads]{llncs}

\usepackage{eccv}

\usepackage{eccvabbrv}

\usepackage{graphicx}
\usepackage{booktabs}

\usepackage{wrapfig}
\usepackage{epsfig}
\usepackage{amsmath,amssymb}
\usepackage{tikz}
\usepackage{comment}
\usepackage{color}
\usepackage{adjustbox}
\usepackage{multirow}
\usepackage{cuted}
\usepackage{mathtools}
\usepackage{floatflt}
\usepackage[boxed, ruled,linesnumbered]{algorithm2e}
\usepackage{algpseudocode}
\usepackage[accsupp]{axessibility}  
\newcommand{\myparagraph}[1]{\vspace{2pt}\noindent{\bf #1}}
\usepackage{hyperref}

\usepackage{orcidlink}

\begin{document}

\title{MEVG: Multi-event Video Generation with Text-to-Video Models} 

\titlerunning{MEVG: Multi-event Video Generation with Text-to-Video Models}

\author{Gyeongrok Oh\inst{1}\orcidlink{0000-0002-4705-0705}
\and
Jaehwan Jeong\inst{1}\orcidlink{0009-0006-1894-5266} 
\and
Sieun Kim\inst{1}\orcidlink{0009-0006-3359-0504}
\and
Wonmin Byeon\inst{2}\orcidlink{0000-0002-4780-4749}
\and \\
Jinkyu Kim\inst{1}\orcidlink{0000-0001-6520-2074}
\and
Sungwoong Kim\inst{1}\orcidlink{0000-0002-2676-9454}
\and
Sangpil Kim\inst{1}\orcidlink{0000-0002-7349-0018}}

\authorrunning{G. Oh et al.}



\institute{$^1$Korea University\quad 
$^2$NVIDIA
}

\maketitle

\begin{abstract}
  We introduce a novel diffusion-based video generation method, generating a video showing multiple events given multiple individual sentences from the user.
Our method does not require a large-scale video dataset since our method uses a pre-trained diffusion-based text-to-video generative model without a fine-tuning process.
Specifically, we propose a last frame-aware diffusion process to preserve visual coherence between consecutive videos where each video consists of different events by initializing the latent and simultaneously adjusting noise in the latent to enhance the motion dynamic in a generated video.
Furthermore, we find that the iterative update of latent vectors by referring to all the preceding frames maintains the global appearance across the frames in a video clip.
To handle dynamic text input for video generation, we utilize a novel prompt generator that transfers course text messages from the user into the multiple optimal prompts for the text-to-video diffusion model.
Extensive experiments and user studies show that our proposed method is superior to other video-generative models in terms of temporal coherency of content and semantics.
Video examples are available on our project page: \url{https://kuai-lab.github.io/eccv2024mevg}.
\end{abstract}

\section{Introduction}
\label{sec:introduction}
Deep generative models in the computer vision community gain a significant spotlight due to their unprecedented performance.
Especially, text-to-image generation model (T2I)~\cite{ramesh2022hierarchical, gu2022vector, rombach2022high, saharia2022photorealistic,ding2022cogview2, kang2023scaling} has successfully produced high-quality images with complex text descriptions.
However, the complexity of spatial-temporal relations for modeling motion dynamics, light conditions, and scene transitions for video generation with deep generative models requires huge computational resources and large-scale text-video paired datasets.
Despite these challenges, recent methods~\cite{blattmann2023align, he2022latent, singer2022make, hong2022cogvideo, khachatryan2023text2video, huang2023free, hong2023large} for text-to-video generation achieve data and cost-efficient training by leveraging the pre-trained text-to-image generative models~\cite{rombach2022high, ding2022cogview2}.
Although spatial-temporal modeling aided by prior knowledge from text-image pairs helps to generate high-quality frames and capture semantically complex descriptions, it falls short in addressing real-world video comprehensively.

In essence, videos in the wild consist of consecutive events with dynamic movements, backgrounds, objects, and viewpoint changes over time.
However, existing approaches mainly generate a video with a single prompt that disregards semantic transitions from event to event and restrictively expresses the entire story when the story consists of multiple events.
Multi-event-based video generation requires three significant criteria: 1) smooth transition between each video clip, 2) semantic alignment between the prompt from the user and the generated video, and 3) ensuring diversity of the content and motion in the video. 

Recently, some studies~\cite{villegas2022phenaki, ge2022long} have made progress in embracing the multiple descriptions that contain chronological sequence events. 
Despite this breakthrough, substantial training efforts are necessary using extensive text-video datasets because they present the additional networks to make multi-prompt video generation.  
The concurrent work~\cite{wang2023gen} leverages the pre-trained text-to-video generation model~(T2V). 
However, the overlapped denoised process on the consecutive distinct prompts induces visual degradation and significant inconsistency between the background and objects.
Moreover, traditional long-term T2V methods~\cite{singer2022make, blattmann2023align, ge2022long, zhang2023show, hong2022cogvideo} hierarchically generate a video by bridging the gap between each keyframe, followed by generating keyframes given a single description. 
This hierarchical video generation process makes it challenging to incorporate the multiple-time variant prompts since it comprehensively generates the global video content in the beginning.

\begin{figure*}[t!]
  \centering
  \includegraphics[width=\linewidth]{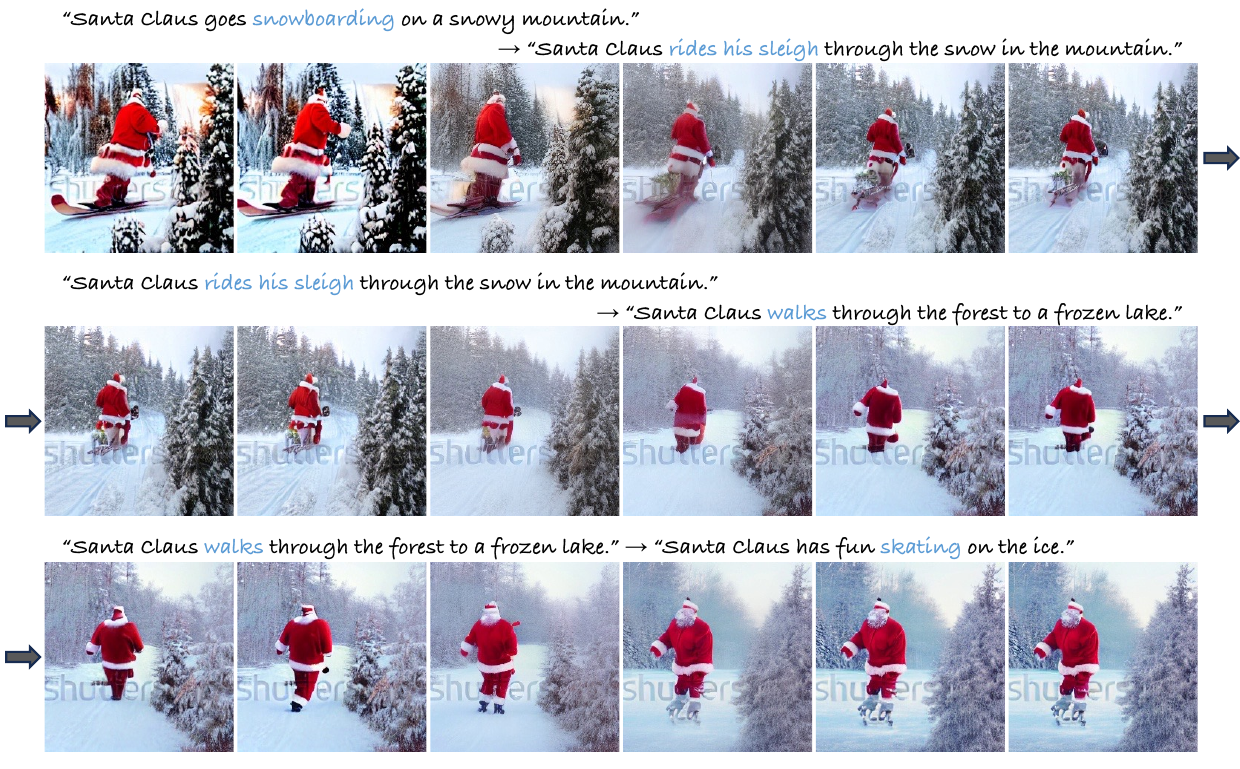}
  \caption{An example of multi-event video generation. MEVG produces impressive output that corresponds to the given prompts and consists of chronologically continuous events.}
  \label{fig:fig_showcase}
\end{figure*}

Our method, multi-event video generation method~(MEVG), is delicately designed to generate a video clip consisting of multiple events without any video data nor fine-tuning process. MEVG successively enforces the temporal coherence between independently generated video clips given multiple event descriptions.
By building upon the publicly released diffusion-based video generation model, we effectively utilize the pre-trained single-prompt T2V generative model\footnote{We used ~\cite{he2022latent}. From our experiments, our proposed method can be applied to any kind of diffusion-based T2V pre-trained model.} to generate complex scenario videos.
Specifically, to preserve the visual coherence between time variable prompts in video generation while producing realistic and diverse motion, we introduce two novel techniques: a last frame-aware latent vector initialization method and a structure-guided sampling strategy for the diffusion-based generative model. First, the last frame-aware latent vector initialization stage includes (\lowercase\expandafter{\romannumeral1}) \textit{dynamic noise}, which diversifies motion across frames, and (\lowercase\expandafter{\romannumeral2}) \textit{last frame-aware inversion}, which guides to generate consistent contents between prompts. \textit{Structure-guided sampling} further improves the visual consistency by progressively updating the latent code during the sampling process.
In addition, to incorporate sequentially structured prompts, we leverage the Large Language Model~(LLM) as a prompt generator. Since a single story has sequentially incorporated events within one sentence, LLM separates a complex story into multiple prompts, each having only one event.

From our extensive experiments and user studies, we demonstrate that our proposed methods generate realistic videos that include three representative types of change: object motion, background, and complex content changes.
Moreover, we examine the effectiveness and legitimacy of each proposed method by conducting ablation studies.
To summarize, our main contributions are as follows:
\begin{itemize}
\item Our proposed diffusion-based video generation method generates a video consisting of multiple events without requiring any training or additional video data.
\item We present a last-frame aware initialization method and dynamic noise adjustment strategy for the latent vector that enhances temporal and semantic consistency between individual videos where each video shows a different event. 
\item We present a novel prompt generator that transforms course text inputs into optimal text instructions for a text-to-video generative model, ensuring coherence of semantic transitions in the generated video.
\item We show that our proposed video generation method outperforms the other zero-shot video generation methods in reflecting multiple events while maintaining visually coherent content for video generation.
\end{itemize}

\section{Related Work}
\label{sec:related}
\myparagraph{Text-to-Video Generation.}
Text-to-video (T2V) generation has shown remarkable progress.
Three primary methodologies are utilized in the field of computer vision.
A Generative Adversarial Network (GAN)~\cite{wang2020g3an,brooks2022generating,yu2022generating,skorokhodov2022stylegan, tian2021good, vondrick2016generating} is a well-known algorithm to generate diverse video from a noise vector utilizing a generator and discriminator. 
Another approach is auto-regressive transformers~\cite{weissenborn2019scaling, hong2022cogvideo, ge2022long, yan2021videogpt, wu2021godiva, wu2022nuwa} that leverage discrete representation to depict the motion dynamics. 
Recently, diffusion-based methods~\cite{ho2022video, singer2022make, ge2023preserve, ho2022imagen, he2022latent, blattmann2023align, zhou2022magicvideo, esser2023structure, voleti2022mcvd, bar2024lumiere, chen2024videocrafter2} have shown significant progress in learning data distribution while iteratively removing noise from the initial gaussian noise.

Long-term video generation has recently been a popular topic in the computer vision community.
Auto-regressive approaches~\cite{ge2022long, hong2022cogvideo, liang2022nuwa} leveraging transformer architecture show plausible results in long-term video generation. However, they require massive training costs and datasets.
Furthermore, although TATS~\cite{ge2022long} and Phenaki~\cite{villegas2022phenaki} can generate videos driven by a sequence of prompts, accumulated errors over time cause drastic changes in video content and visual quality degradation due to the auto-regressive property.
Several works~\cite{singer2022make, ge2023preserve, blattmann2023align, zhang2023show, voleti2022mcvd} based on the diffusion model leverage temporal interpolation networks and masked strategies for generating smoother videos. 
VidRD~\cite{gu2023reuse} directly utilizes the previous initial latent code to expand the video.

Note that most previous works focused on video generation from a single prompt or an event. However, in this work, we tackle the multi-event video generation task, which consists of consecutive events in a long-term video.
Animate-A-Story~\cite{he2023animate} utilizes abundant real-world video corresponding to each story for natural motion.
Moreover, SEINE~\cite{chen2023seine} focuses on the transition from short to long video generation models by utilizing millions of datasets.
Gen-L-Video~\cite{wang2023gen}, another approach, uses overlapped frames between two successive prompts. Although this strategy makes the outcomes more realistic, undesirable contents occur due to the overlapping denoising process.

\begin{figure*}[t!]
  \centering
  \includegraphics[width=\linewidth]{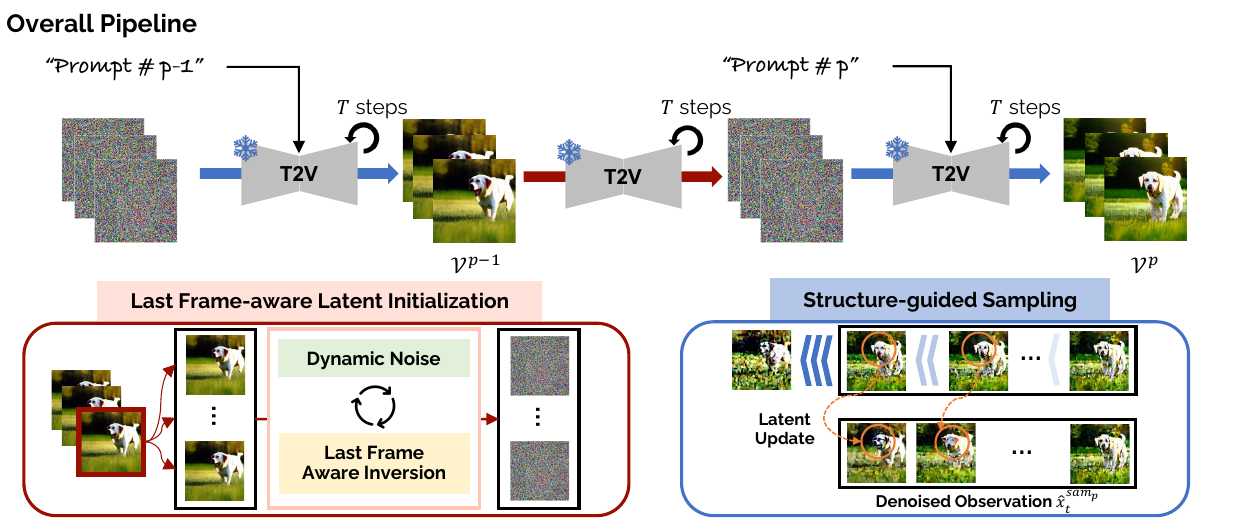}
  \caption{MEVG synthesizes the consecutive video clips corresponding to distinct prompts. The overall pipeline comprises two major components: last frame-aware latent initialization and structure-guided sampling. First, in the last frame-aware latent initialization, the pre-trained text-to-video generation model adopts the repeated frame as an input to invert into the initial latent code with two novel techniques: \textit{dynamic noise} and \textit{last frame-aware inversion}.
  Second, \textit{structure-guided sampling} enforces continuity within a video clip by updating the latent code.
  }
  \label{fig:overview}
\end{figure*}

\myparagraph{Zero-shot approach.}
FateZero~\cite{qi2023fatezero} and INFUSION~\cite{khandelwal2023infusion} edit video by leveraging the pre-trained image diffusion models while ensuring temporal consistency. 
These methods utilize attention maps and spatial features to preserve the structure and temporal coherence over the frame. 
In the generation field, Text2Video-Zero~\cite{khachatryan2023text2video} synthesizes the video to keep the global structure across the sequence of frames without video data while encoding the motion dynamics to provide diverse movement. 
To encode the movement, latent codes enclose the motion dynamics with direction parameters.
Free-bloom~\cite{huang2023free} and DirecT2V~\cite{hong2023large} are distinct approaches employing the text-to-image models while sharing a similar conceptual framework.
Both utilize the Large Language Model~(LLM) in order to maintain the semantic information for each generated frame. 
DirecT2V leverages self-attention to preserve the appearance of the video; in addition, Free-bloom leverages joint distribution to sample the initial code for consistent frames.

Inspired by these approaches, we leverage a pre-trained text-to-video~(T2V) generation model to extend a short, monotonous video into an exciting video containing variable events. To pursue the naturalness of results, challenges exist in maintaining visual coherence and guaranteeing diversity.
Therefore, we adjust the latent code near the preceding video and grant dynamic changes through gradual perturbation. 
\section{Method}
\label{sec:methodology}
We propose a novel pipeline to generate a temporally and semantically coherent video conditioned on multiple event-based prompts.
Specifically, our goal is to generate multiple video clips without disturbing the natural flow and recurrent video pattern across the semantic transitions in the given prompts. 
In this section, we first provide the introductory diffusion model that is the basis for our research and an overview of our proposed pipeline~(see Sec.~\ref{sec:preliminary} and Sec.~\ref{sec:pipeline}). 
Next, we present the technical details of our two main components: 
(\lowercase\expandafter{\romannumeral1}) \textit{last frame-aware latent initialization}, (\lowercase\expandafter{\romannumeral2}) \textit{structure-guided sampling}, in Sec.~\ref{sec:inversion} and Sec.~\ref{sec:sampling}.  
Finally, we introduce the prompt generator, harnessing the powerful ability of Large Language Model~(LLM) to handle a complex story containing multiple meaningful events~(Sec.~\ref{sec:prompt}).

\subsection{Preliminaries}
\label{sec:preliminary}
\myparagraph{DDPM.}
The diffusion probabilistic model~\cite{ho2020denoising} has two components: a forward diffusion process and a backward diffusion process. In the forward process, data distribution transforms into noise distribution by adding noise iteratively. For every timestep $t$, noise $\epsilon\sim N(0,I)$ diffract the original data $x$ utilizing the variance schedule $\beta_t$ as follows:
\begin{equation}
    x_t = \sqrt{\bar{\alpha}_t}x + \sqrt{1-\bar{\alpha}_t}\epsilon,
\end{equation}
where $\alpha_t=1-\beta_t$ and $\bar{\alpha}_t = \prod_{i=0}^t\alpha_i$.
In training time, the diffusion model predicts noise during every step to reconstruct the original data distribution. This reverse process $q(x_{t-1}|x_{t})$ is parameterized as follows:
\begin{equation}
    p_{\theta}(x_{t-1}|x_{t}) := \mathcal{N}{(x_{t-1};\mu_{\theta}(x_{t}, t),\Sigma_{\theta}(x_{t}, t)}). 
    \label{eq:reverse_process}
\end{equation}

\myparagraph{DDIM.} DDIM~\cite{song2020denoising} is a variant of DDPM that leverages the non-Markovian manner instead of the Markov chain. 
DDIM sampling strategy makes the diffusion process to be deterministic, which can be written as follows:

\begin{equation}
\begin{aligned}
    x_{t-1}= \sqrt{\bar{\alpha}_{t-1}}\underbrace{\bigg{(}{{x_{t}-\sqrt{1-\bar{\alpha}_t}\epsilon_\theta(x_t,t)}\over{\sqrt{\bar{\alpha}_t}}}\bigg{)}}_{\hat{x}_{t}} \\
    +\sqrt{1-\bar{\alpha}_{t-1}-\sigma_t^2}\underbrace{\cdot \epsilon_\theta(x_t,t)}_{\epsilon_t}
    -\sigma_t n_t\;,
    \label{eq:ddim}
    \end{aligned}
\end{equation}
where $\hat{x}_{t}$ indicates the denoised observation of $x_0$ at each diffusion step $t$, $\epsilon_t$ denotes the predicted noise at each time step $t$ and $\sigma$ controls whether the model is stochastic or deterministic.
In this paper, we use the modified DDIM Inversion to strengthen the naturalness of the video, maintaining overall visual consistency despite the semantic changes along the temporal axis in the given prompts.

\subsection{MEVG pipeline}
\label{sec:pipeline}

\begin{wrapfigure}[18]{hr!}{0.55\textwidth}
  \centering
  \includegraphics[width=\linewidth]{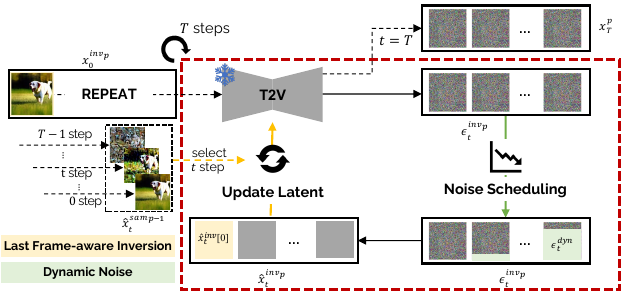}
  \caption{\textbf{Last Frame-aware Latent Initialization} Initial latent code is crucial for maintaining global geometric structure. We apply two techniques performing different roles: (\lowercase\expandafter{\romannumeral1}) \textit{dynamic noise} tailors flexibility differentially across each frame, and (\lowercase\expandafter{\romannumeral2}) \textit{last frame-aware inversion} restricts the model to minimize the divergence of the entire frames from the content of the preceding video clip.}
  \label{fig:LFALI}
\end{wrapfigure}

We outline our proposed MEVG that utilizes the former video clip to generate subsequent video clips considering the given prompts (see Fig. \ref{fig:overview}).
Our method is built upon the latent video diffusion model~\cite{he2022latent}, which leverages the low-dimensional latent space  $x\in\mathbb{R}^{F\times c\times h\times w}$, where $c,\;h,$ and $\;w$ denote latent space dimension and $F$ indicates the total number of frames. 
The video output $\mathcal{V}\in\mathbb{R}^{F\times3\times H\times W}$ are obtained by passing the latent code $x$ through the decoder $\mathcal{D}$, where $H\times W$ is the resolution of the frame.

To get the final result $\textrm{V}=\{\mathcal{V}^{p}\}_{p=0}^{\mathcal{P}-1}$, where $\mathcal{P}$ denotes the number of given prompts, we first sample the video conditioned on the first prompt.
An initial video clip is created by Gaussian distribution $x_T\sim\mathcal{N}(0,I)$ as well as a static image as conditional guidance to capture the essential visual content corresponding to user intention.
After generating the initial video clip, we extend a preceding video in accordance with the semantic context of the subsequent prompt, driven by two major elements: \textit{last frame-aware latent initialization} and \textit{structure-guided sampling}.
\textit{Last frame-aware latent initialization} is the initialization process of the noise latent that helps to preserve the spatial information while generating more diverse contents. \textit{Structure-guided sampling} then enforces motion consistency between frames during the backward diffusion process.

\subsection{Last Frame-aware Latent Initialization}
\label{sec:inversion}

The inversion technique, which reconstructs the initial latent code from the visual input~(e.g., image or video), is used in real-world applications~\cite{xing2023inversion, nguyen2023visual, dong2023prompt}, where accurate spatial layout or visual content reconstruction is important.
Video generation should have visual coherence across the entire sequence of frames while adapting the movement of objects and the background transition.
To achieve this goal, we aim to find the optimal latent code that helps preserve global coherence between the generated videos for the previous and next prompts and maintains an ability to adapt to the changes.
However, the existing approaches~\cite{gu2023reuse, wang2023gen} present repetitive video patterns~(e.g., similar camera movement and object position) and awkward scene transition caused by overlapped content in a single frame.

\begin{wrapfigure}[23]{R}{0.55\textwidth}
\scalebox{0.73}{
    \begin{minipage}{0.73\textwidth}
\SetKwComment{Comment}{\# }{}
\RestyleAlgo{ruled}
\setlength{\textfloatsep}{8pt}
\begin{algorithm}[H]
	\setlength{\belowcaptionskip}{-20pt}
	\caption{\scriptsize Last Frame-aware Latent Initialization}
	\label{alg:inversion}  
        \Indm 
        \KwIn{
            latent code of the the previous prompt $x_0^{p-1}$, denoised observation of the last frame from the previous prompt $\{\hat{x}_{t}^{sam_{p-1}}[-1]\}_{t=0}^{T-1}$, noise scheduling function $\mathcal{F}(\cdot$), and pre-trained \text{T2V} model \textbf{\text{T2V}}($\cdot$)
        }
	\KwResult{
            Initial latent code $x_T^{p}$ of next prompt $p$
        }
        \tcp*[h]{\small $T=$ Number of diffusion steps}
        
        \tcp*[h]{\small $N=$ Number of frames}
        
        \tcp*[h]{\small $p=$ Index of next prompt}

        \Indp
        $x_0^{inv_p}\leftarrow\text{REPEAT}(x_{0}^{^{p-1}}[-1])$
        
        \For{$t$ in $0,...,T-1$}{
            $\epsilon_{t}^{inv_p} \leftarrow \textbf{T2V}(x_t^{inv_p}, t)$

            \tcp{\textcolor{blue}{Dynamic Noise}}

            \For{$n$ in $0,...,\textrm{N}-1$}{
                $\kappa_n \leftarrow \mathcal{F}(n)$
                
                $\epsilon^{dyn}_{t}\sim\mathcal{N}(0, {1\over{1+\kappa_n^{2}}}\mathrm{I})$
            
                $\epsilon^{inv_p}_{t}[n]\leftarrow{\kappa_n\over{\sqrt{1+\kappa_n^2}}}\epsilon^{inv_p}_{t}[n]+\epsilon^{dyn}_{t}$
                }
                
            $\hat{x}_{t}^{inv_p}\leftarrow (x_t^{inv_p}-\sqrt{1-\bar{\alpha}_t}\epsilon_{t}^{inv_p})/\sqrt{\bar{\alpha}_t}$

            \tcp{\textcolor{blue}{Last Frame-aware Inversion}}
            
            $\mathcal{L}_{\text{LFAI}} = ||\hat{x}_{t}^{sam_{p-1}}[-1]-\hat{x}_{t}^{inv_{p}}[0]||_{2}^{2}$

            $\hat{x}_{t}^{inv_p}\leftarrow \hat{x}_{t}^{inv_p}-\delta_{\text{LFAI}}\nabla_{\hat{x}_{t}}\mathcal{L}_{\text{LFAI}}$

            $x_{t+1}^{inv_p}\leftarrow \sqrt{\bar{\alpha}_{t+1}}\hat{x}_{t}^{inv_p} + \sqrt{1-\bar{\alpha}_{t+1}}\epsilon_{t}^{inv_p}$
        }        
        
        $x_T^{p} = x_{T}^{inv_p}$
\end{algorithm}
    \end{minipage}}
\end{wrapfigure}

To solve these challenges, we reuse the generated video (essentially the last frame) from the previous prompt to generate the frames for the new prompt. 
Basically, the last frame of the previously generated video is copied over the entire sequence as an initial conditioning input. 
We then propose dynamic noise to enforce the diversity of the generated video.
This process preserves the overall visual contents, such as an object and background across the video, and also improves the generation diversity.

\myparagraph{Dynamic Noise.}
To generate diverse motion of the object and smooth transition of the background given prompts, we add the video noise prior similar to \cite{ge2023preserve}.
Essentially, the noise vector  $\epsilon_{t}^{dyn}\sim\mathcal{N}(0, {1\over{1+\kappa^2}}\mathrm{I})$ is added to the predicted noise $\epsilon_{t}^{inv_p}$ during inversion stage for next prompt $p$.
$\kappa$ regulates the dynamics and variability of the frames within a single video segment; $\kappa\rightarrow 0$ increases video variations.

Since the beginning of the new video should be similar to the preceding video clip, and then more changes occur toward the end of the video, we design a noise scheduling function $\mathcal{F}=\exp(-x)$ that monotonically decreases the $\kappa$.
$\kappa_n$ corresponding to the frame index $n$ is determined by:
\begin{equation}
    \kappa_n = \mathcal{F}(n),\quad 0 \leq n<N,
\end{equation}
where $N$ is the total number of frames within one video clip.
Finally, the predicted noise $\epsilon^{inv_p}_t$ is obtained as follows:
\begin{equation}
    \epsilon_t^{inv_p}[n] = {\kappa_n\over{\sqrt{1+\kappa_n^2}}}\epsilon_{t}^{inv_p}[n]+\epsilon_{t}^{dyn}, \quad 0 \leq n<N,
\end{equation}
where $[\cdot]$ denotes the index of frame.

\myparagraph{Last Frame-aware Inversion.}
\label{sec:LFAI}
While \textit{dynamic noise} helps to generate diverse video contents, they cause a temporal inconsistency problem between individual video clips conditioned on consecutive prompts.
\textit{Last frame-aware inversion} coerces to maintain a visual correlation between different video clips guided by the denoised observation $\hat{x}_{t}$. 
Since the denoised observation $\hat{x}_{t}$ is the predicted noise-free latent at diffusion step $t$~(see Eq.~\ref{eq:ddim}), it contains a sketchy spatial layout and video context.
We regularize the initial frame of the current video clip $\hat{x}_{t}^{inv_p}[0]$ using the denoised observation of the last frame $\hat{x}^{sam_{p-1}}_{t}[-1]$ from the previous clip. This process ensures the visual consistency between two video clips.
We minimize the objective $\mathcal{L}_{\text{LFAI}}$ using L2 loss as follows:

\begin{equation}
    \mathcal{L}_{\text{LFAI}} = ||\hat{x}_{t}^{sam_{p-1}}[-1]-\hat{x}_{t}^{inv_p}[0]||_{2}^{2}.
\end{equation}
It basically aligns the denoised observations between the sampling process for the previous prompt and the inversion process for the next prompt at each diffusion step $t$.
After all, we update the $\hat{x}_{t}^{inv_p}$, the denoised observation during the inversion procedure, along the direction that minimizes the $\mathcal{L}_{\text{LFAI}}$ and $\delta_{\text{LFAI}}$ controls the guidance strength.

Consequently, through this procedure, we maintain the flexibility allowed by the \textit{dynamic noise} and regularize the overall visual content by employing the denoised observation $\hat{x}_{t}$.
We present the procedure of \textit{last frame-aware latent initialization} in Alg.~\ref{alg:inversion} to facilitate understanding, and the overall procedure are illustrated in Fig.~\ref{fig:LFALI}.

\subsection{Structure-guided Sampling}
\label{sec:sampling}

The video clip is generated for the next prompt using the initial latent $x^p_T$ produced at the previous step.
Although the video clip for the new prompt should preserve the appearance of the previous video clip by using the last frame-aware initial latent, undesirable variation in scene texture and object placement often occurs due to the stochastic nature of the sampling process.
To improve the visual consistency within a video clip, we progressively update the predicted original $\hat{x}_{t}^{sam_p}$ of the current video clip in the sampling process.
Specifically, we formulate the objective as follows:
\begin{equation}
\label{eq:sgs_object}
 \mathcal{L}_{\text{SGS}} = ||\hat{x}_{t}^{sam_p}[1:n]-\hat{x}_{t}^{sam_p}[:n-1]||_{2}^{2},   
\end{equation}
where $n\in\{1,...,\textrm{N}\}$. 
Note that for the first frame ($n=0$), we compute $\mathcal{L}_{SGS}$ using the denoised observation of the last frame from the previous prompt $\{\hat{x}_{t}^{sam_{p-1}}[-1]\}_{t=0}^{T-1}$.
Finally, we update $\hat{x}_{t}^{sam_p}$ as follows:
\begin{equation}
\label{eq:sgs_total}
    \hat{x}_{t}^{sam_p}\leftarrow \hat{x}_{t}^{sam_p}-\delta_{\text{SGS}}\nabla_{\hat{x}_{t}}\mathcal{L}_{\text{SGS}},
\end{equation}
where $\delta_{\text{SGS}}$ is responsible for guidance scale.
Eq.~\ref{eq:sgs_object} and Eq.~\ref{eq:sgs_total} are iteratively conducted frame-by-frame at each diffusion step. Guidance on the denoised observation leads to a similar global geometric structure between frames within a single video clip.

\subsection{Prompt Generator}
\label{sec:prompt}
In real-world scenarios, multiple sequential events can be described in one sentence or paragraph. For instance,  ``\textit{The dog runs across the wide field, then comes to a halt, yawns softly, and lies down."}.
However, existing long-video generation models~\cite{ge2022long, hong2022cogvideo, he2022latent} are designed to generate only a single event.
Therefore, the generated video does not reflect the entire text when the prompt contains multiple events.   
To address this issue, the Large Language Model~(LLM) has been utilized to generate an appropriate input for the pre-trained T2V models.
We introduce a prompt generator to segment the comprehensive description into the prescribed textual format.
We put the exemplar and guidelines in the supplementary materials.

\section{Experiments}
\label{sec:experiment}

\begin{figure*}[t]
\includegraphics[width=\textwidth]{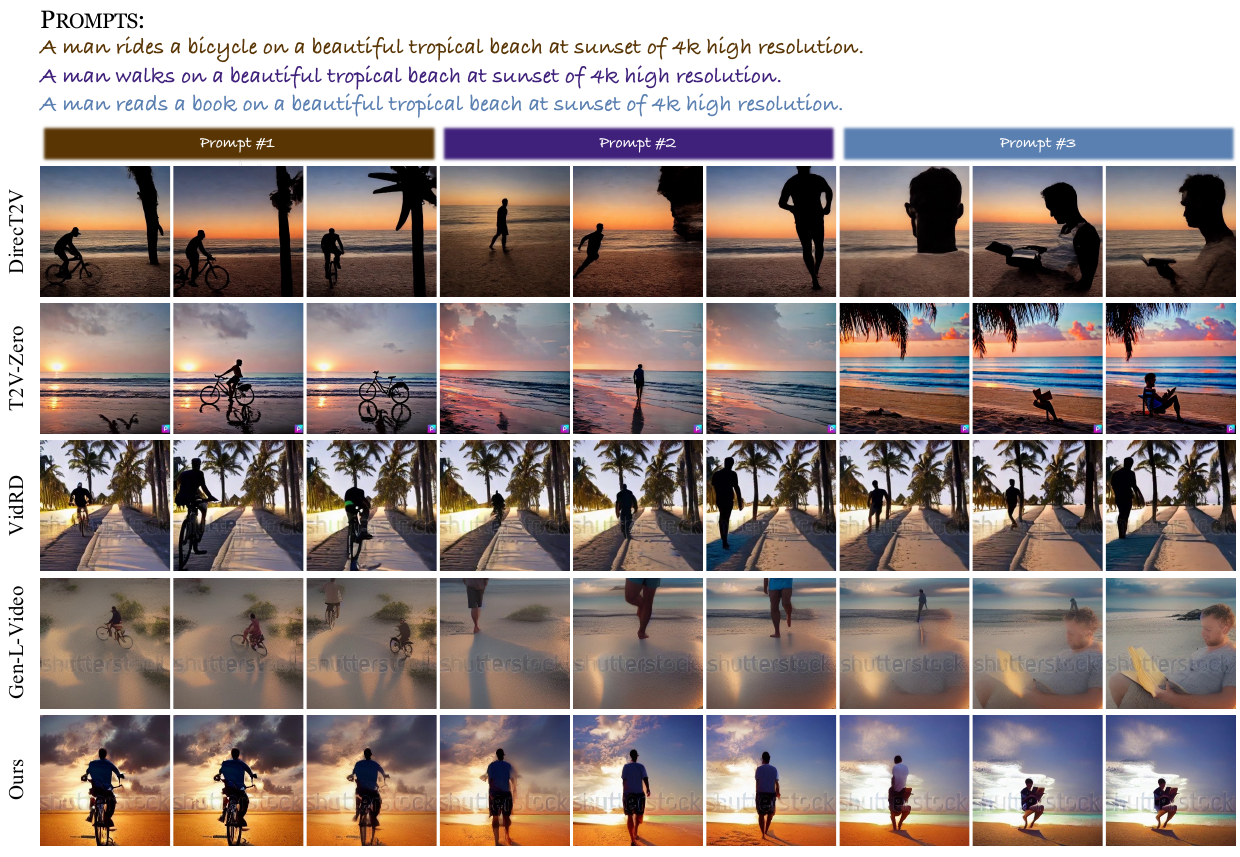}
\captionof{figure}
{Generation results on given prompts by our method and baseline models. T2V-Zero and DirecT2V build upon the T2I pre-trained model. In contrast, VidRD and Gen-L-Video leverage the same foundation model utilized in our experiments.}
\label{fig:fig_qualitative}
\end{figure*}

\subsection{Implementation Details}
We directly leverage the released pre-trained text-to-video generation model~\cite{he2022latent} to generate a multi-text conditioned video.
In our experiments, we generate multi-text conditioned video, and each video clip consists of 16 frames with 256$\times$256 resolution. 
Moreover, we employ  ChatGPT~\cite{ouyang2022training}, which is a Large Language Model, to separate the complex scenarios into individual prompts that comprise the sequence of events. 
We set each guidance weight $\delta_\text{LFAI}$ and $\delta_\text{SGS}$ to 1000 and 7 in our experiments.
All experiments are performed on a single NVIDIA GeForce RTX 3090.

\begin{table*}[t!]
\centering
\caption{Compared with baseline methods in terms of two primary categories: automatic metric and human evaluation. Note that we use {\bf{bold}} to highlight the best scores, and \underline{underline} indicates the second-best scores.}
\label{table:Quantitative_baseline}
\begin{adjustbox}{width=\textwidth,center}{
\begin{tabular}{lccccccr}
\toprule
& \multicolumn{2}{c}{Automatic Metric} & \multicolumn{4}{c}{Human Evaluation} \\
\cmidrule(l{1mm}r{1mm}){2-3} 
\cmidrule(l{1mm}r{1mm}){4-7} 
Method  &CLIP-Text $\uparrow$ & CLIP-Image $\uparrow$ &  Temporal $\uparrow$ & Semantic $\uparrow$  & Realism $\uparrow$ & Preference $\uparrow$ \\
\midrule
\textcolor{gray}{T2V-Zero~\cite{khachatryan2023text2video}}  & \textcolor{gray}{\textbf{0.322}} & \textcolor{gray}{0.808} &  \textcolor{gray}{\underline{3.61}}  &  \textcolor{gray}{\underline{3.59}} & \textcolor{gray}{3.45} &  \textcolor{gray}{\underline{3.47}} \\
\textcolor{gray}{DirecT2V}~\cite{hong2023large}  & \textcolor{gray}{0.301} & \textcolor{gray}{0.898} &  \textcolor{gray}{2.96} &  \textcolor{gray}{3.04} &  \textcolor{gray}{3.01} &  \textcolor{gray}{3.30} \\
Gen-L-Video~\cite{wang2023gen}  & 0.308 & \underline{0.953} & 3.35 & 3.38 & 3.37 & 3.05\\
VidRD~\cite{gu2023reuse} & 0.287 & 0.951 & 3.40 & 3.43 & \underline{3.56} & 3.14\\
Ours & \underline{0.309} & \textbf{0.957} & \textbf{3.82} & \textbf{3.71} & \textbf{3.68}  & \textbf{3.68} \\
\bottomrule
\end{tabular}}
\end{adjustbox}
\end{table*}

\subsection{Qualitative Results}

We provide qualitative comparisons along with other recent multi-prompts video generation methods~\cite{wang2023gen, gu2023reuse}, including zero-shot video generation methods~\cite{hong2023large, khachatryan2023text2video} which leveraging frame-level descriptions.
In the supplementary material and the project page, we provide additional videos for frame-level and video-level comparison to show more qualitative examples.
As shown in Fig.~\ref{fig:fig_qualitative}, our proposed method achieves better video quality, especially in two points: naturalness and temporal coherence.
First, compared with T2V-Zero~\cite{khachatryan2023text2video} and DirecT2V~\cite{hong2023large}, we observe that only leveraging the image-based model fails to generate reasonable video flow in terms of naturalness; thus, utilizing the video-based approaches is necessary.
Second, generated videos by our method show a strong visual relation between each video segment without the recurrent video pattern.
To be more specific, visual examples of VidRD~\cite{gu2023reuse} exhibit recurrent video patterns between two distinct video clips; e.g., the movement pattern of a man within each video segment mirrors the previous one.
Additionally, although Gen-L-Video~\cite{wang2023gen} generates more diverse movement, they can not preserve the structure coherence of objects, and the background is not stable across the entire frame.
On the other hand, our method not only smoothly bridges the gap between two individual video clips but also maintains the overall temporal coherence of the content.

\subsection{Quantitative Results}

\noindent
\begin{wrapfigure}{r}{0.5\textwidth}
\includegraphics[width=\linewidth]{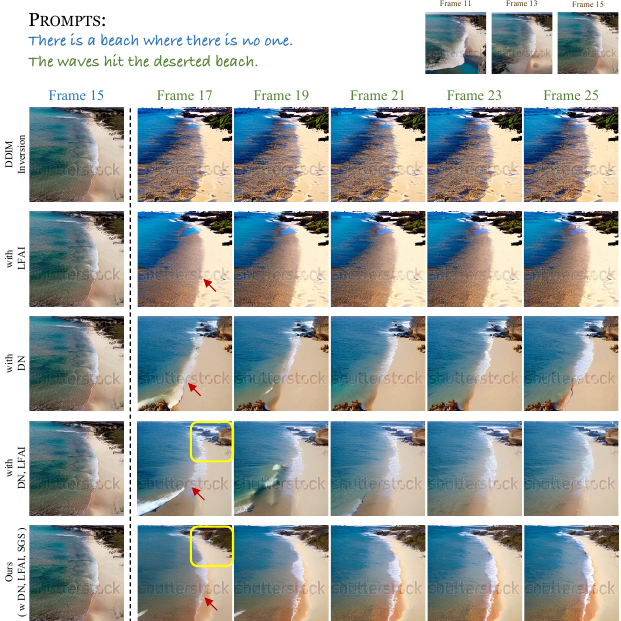}
  \caption{Generated video clips \textit{with} and \textit{without} proposed our modules. The red arrow and yellow box highlight the visual changes between distinct video clips.}
\label{fig:ablation_study_module}
\end{wrapfigure}

\myparagraph{Automatic Metrics.}
We report the CLIP-Text score~\cite{radford2021learning, hessel2021clipscore} that represents the alignment between given prompts and outputs, and CLIP-Image score~\cite{esser2023structure, qi2023fatezero, Ceylan_2023_ICCV, wu2023tune} that shows the similarity between two consecutive frames. 
We measure the metrics over the 30 scenarios, each consisting of multiple prompts. 
For a fair evaluation, we randomly sampled 20 videos per scenario.
As shown in Tab.~\ref{table:Quantitative_baseline}, our method generally outperforms the other state-of-the-art methods.
Among the baseline, the generated video of Text2Video-Zero~(T2V-Zero) aligns well with the semantics of given prompts as this approach yields a single frame corresponding to a prompt of the same video segments.
However, they fail to generate temporally coherent video.
In contrast, DirecT2V~\cite{hong2023large}, which utilizes the frame-specific descriptions sharing high-level stories, shows a higher CLIP-Image score, whereas we observe a decrease in performance for the CLIP-Text score.
Comparison with text-to-video-based approaches exhibits a relatively low difference in all metrics due to the common foundation model~\cite{he2022latent}.
The CLIP-Text score of VidRD~\cite{gu2023reuse} is notably lower than the baselines. The visual content is substantially maintained without a significant performance drop in CLIP-Image score since the VidRD directly utilizes the latent code of the previous video clip with minimal deviations from the sampling step. 
Gen-L-Video~\cite{wang2023gen} performs well in capturing the meaning of the prompts. However, the global content variations during the sampling process caused by overlapping prompts lead to a decrease in similarity between consecutive frames.

\myparagraph{Human Evaluation.}
We recruited 100 participants through Amazon Mechanical Turk (AMT) to evaluate five models: T2V-Zero~\cite{khachatryan2023text2video}, DirecT2V~\cite{hong2023large}, Gen-L-Video~\cite{wang2023gen}, VidRD~\cite{gu2023reuse}, and our method.
We employ a Likert scale ranging from 1 (low quality) to 5 (high quality).
Participants score each method considering temporal consistency, semantic alignment, realism, and preference over 30 videos generated by different scenarios.
As clearly indicated in Tab.~\ref{table:Quantitative_baseline}, generated videos from our method significantly outperform other state-of-the-art approaches in all four criteria, regardless of individual frame quality.
In particular, based on human evaluation results, we observe that preserving the identity of the object and background is crucial for human preference. 
When compared with text-to-video-based methods, temporal inconsistency between each video clip caused by the semantic transition of given prompts results in lower human evaluation scores in spite of the same foundation model as ours.

\subsection{Ablation Studies}
\label{sec:ablation}

\myparagraph{Effectiveness of Proposed Methods.}
We qualitatively show the effectiveness of \textit{last frame-aware inversion}~(LFAI), \textit{dynamic noise} (DN), and \textit{structure-guided sampling} (SGS), as shown in Fig.~\ref{fig:ablation_study_module}.
We utilize the basic inversion strategy~\cite{song2020denoising} as a base model.
Although basic DDIM inversion somewhat preserves the overall structure of visual content between each video clip, the detailed texture and background show severe changes.
DN model relaxes this problem but shows the disconnection between two video clips since the DN module gives flexibility to the model by using the i.i.d noise in the inversion procedure.
Combining two modules, LFAI and DN, generates natural-looking videos since LFAI module preserves the structure of the content in the previous frame.
However, we find that the stochastic characteristics of the sampling process introduce slight fluctuations in the videos, and iterative update of latent code~(SGS) during the sampling process is beneficial in enhancing the realism of video content, as shown in the fifth row at Fig.~\ref{fig:ablation_study_module}. Moreover, as reflected in Tab.~\ref{tab:ablation_user}, we provide an additional human evaluation to validate the impact of our proposed module by human judges.

\begin{wraptable}[13]{h}{0.5\textwidth}
        \caption{We report user study results on ablation studies using four different criteria: Temporal, Semantics, Realism, and Preference. Note that we use \textbf{bold} to highlight the best scores, and \underline{underline} indicates the second-best scores.}
	\label{tab:ablation_user}
        \resizebox{\linewidth}{!}{%
        \begin{tabular}{@{}cccccccc@{}} \toprule
            & \multicolumn{3}{c}{Method} & \multicolumn{4}{c}{Human Evaluation}  \\\cmidrule{5-8}
            & LFAI & DN & SGS & Temporal $\uparrow$ & Semantics $\uparrow$ & Realism $\uparrow$ & Preference $\uparrow$ \\\midrule
            & - & - & - & 3.42 & 3.40 & 3.45 & 2.89 \\
            & $\checkmark$ & - & - & 3.48 & 3.40 & 3.50 & \underline{2.93} \\
            & - & $\checkmark$ & - & 3.53 & 3.46 & \underline{3.63} & 2.51 \\
            & $\checkmark$ & $\checkmark$ & - & \underline{3.61} & \textbf{3.58} & 3.58 & 2.78 \\
            & $\checkmark$ & $\checkmark$ & $\checkmark$ & \textbf{3.70} & \underline{3.47} & \textbf{3.69} & \textbf{3.27} \\
           \bottomrule
        \end{tabular}}
\end{wraptable}

\begin{figure}[t!]
\centering
\begin{minipage}[b]{.48\textwidth}
  \centering
  \includegraphics[width=\linewidth]{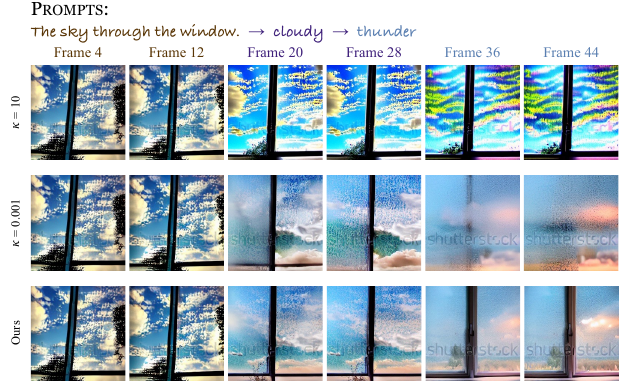}
  \caption{Ablation study for validating the noise schedule. Note that the first and second-row leverage constant value over the entire frames.}
  \label{fig:ablation_study_keppa}
\end{minipage}\hfill
\begin{minipage}[b]{.48\textwidth}
  \centering
  \includegraphics[width=\linewidth]{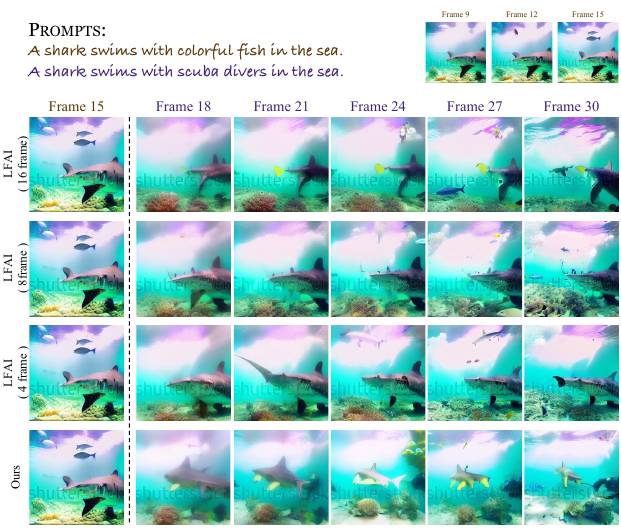}
  \caption{Effectiveness of adjusting the number of influenced frames. }
  \label{fig:ablation_study_lfai}
\end{minipage}
\end{figure}

\myparagraph{Analysis on Dynamic Noise.}
Fig.~\ref{fig:ablation_study_keppa} indicates that $\kappa$ controls the flexibility of the frame sequence. We modify the noise scheduling function $\mathcal{F}$ into the static value to validate the effectiveness of our method. When $\kappa$ applies to the entire frames as a smaller value, we figure out that the latter frames can not preserve the geometric structure. However, frozen video is observed when $kappa$ is set to a high value. 
As a result, we achieve the smooth transition and flexibility between consecutive video clips by adopting the scheduling function $\mathcal{F}$, which decrease steadily.

\myparagraph{Analysis on Last Frame-aware Inversion.} 
We introduce the \textit{last frame-aware inversion} to prevent the visual inconsistency in terms of object spatial location and scene texture driven by the Dynamic Noise. 
In the LFAI process, we guide the first frame to adjust the initial latent code that is correlated to the geometric structure of the previous video clip.
Here, we explore the influence of the number of frames that offer guidance by the last frame of the previous video clip.
As shown in Fig.~\ref{fig:ablation_study_lfai}, we observe that only the first frame is sufficient to maintain the visual structure at the beginning of the frames. On the contrary, the increase in the number of affected frames makes the stationary movement in objects; e.g., the shark only moves on the right side. 
Conversely, the restriction of the affected frame as only a single one gives increased flexibility to the subsequent frames. This flexibility enhances their ability to effectively convey the meaning of the subsequent prompts and generate a diverse range of movement.

\subsection{Applications}
\begin{figure*}[t!]
  \centering
\includegraphics[width=\textwidth]{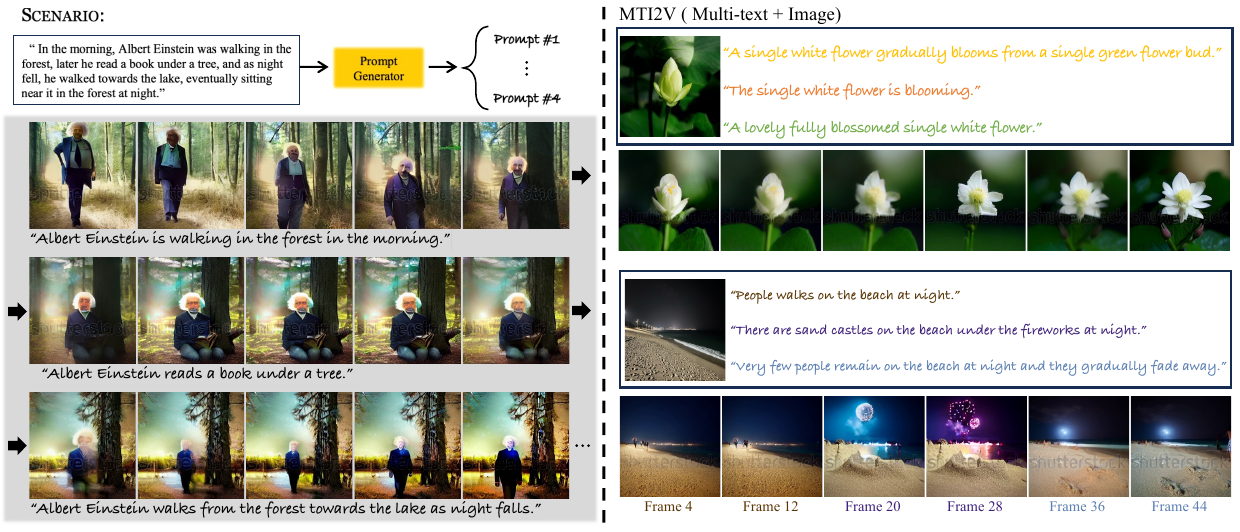}
  \caption{\textbf{(Left)} Example that leverages the Large Language Model~(LLM). Given the complex scenario, our prompt generator split into each individual prompt using the pre-defined instructions.~\textbf{(Right)} Example of our results conditioned on multiple prompts and given image.}
  \label{fig:application}
\end{figure*}

\myparagraph{Video Generation with Large Language Model~(LLM).} In real-world scenarios, more intricate descriptions are generally used, which have the time-variant events in a single narrative.
$\textit{Prompt generator}$~(see Sec.~\ref{sec:prompt}) to separate into the individual prompts for handling the consecutive events.
As shown in Fig.~\ref{fig:application}~(left), the visual examples indicate the entire frame ensures temporal consistency while reflecting the overall storyline.

\myparagraph{Image and Multi-event-based Video Generation.}
Our proposed MEVG is capable of generating video with a given image and multi-text, multi-text-image-to-video generation~(MTI2V). For generating video, we first encode the seeding image using the encoder into the latent vector and duplicate it as the number of frames. Then, we follow MEVG pipeline. Fig.~\ref{fig:application}~(right) demonstrates that the generated video successfully preserves the visual appearance and structure of the object in the reference image and shows temporal coherence along the given prompts.

\section{Conclusion}
\label{sec:conclusion}

We introduced a novel method that generates multi-text-based videos by taking temporally consecutive descriptions.
Specifically, we propose two techniques, \textit{last frame-aware latent initialization} and \textit{structure-guided sampling}, to preserve the visual and temporal consistency in the generated video.
Our proposed method can generate much more natural and temporally coherent videos than the other state-of-the-art methods with qualitative and quantitative results.
Our pipeline also handles a single story containing time-variant events by utilizing the Large Language Model~(LLM). 
In addition, our proposed method can generate videos conditioned on both the multi-prompts and a reference image and can be used in various applications.

\myparagraph{Limitation and Future Work}
Although our proposed method yields promising outcomes in preserving visual consistency and generation diversity over the distinct prompts, there exists potential for future works as listed:
1) Our method requires a certain text format as our model inherits the characteristics of a pre-trained single-prompt video generator
and 2) The absence of benchmark datasets for multi-text video generation makes it hard to conduct a quantitative evaluation.
Video generation with diverse input conditions and curating multi-prompts video datasets are promising future directions.


\section*{Acknowledgements}
This research was supported by Culture, Sports and Tourism R\&D Program through the Korea Creative Content Agency grant funded by the Ministry of Culture, Sports and Tourism in 2024((International Collaborative Research and Global Talent Development for the Development of Copyright Management and Protection Technologies for Generative AI, RS-2024-00345025, 25\%),(Research on neural watermark technology for copyright protection of generative AI 3D content, RS-2024-00348469, 25\%),(Development of sketch-based semantic 3D modeling technology for creating user-centric Metaverse content spaces for indoor spaces, RS-2023-00227409, 10\%))), 
Institute of Information \& communications Technology Planning \& Evaluation (IITP) grant funded by the Korea government(MSIT)(RS-2019-II190079, 10\%),
Artificial intelligence industrial convergence cluster development project funded by the Ministry of Science and ICT(MSIT, Korea)\& Gwangju Metropolitan City(contribution rate: 15\%), and IITP under the Leading Generative AI Human Resources Development(IITP-2024-RS-2024-00397085, 15\%) grant funded by the Korea government(MSIT).

%
%
\bibliographystyle{splncs04}
\bibliography{main}

\title{Supplemental Material to:\\MEVG: Multi-event Video Generation with Text-to-Video Models} 

\titlerunning{MEVG: Multi-event Video Generation with Text-to-Video Models}

\author{}
\institute{}

\authorrunning{G. Oh et al.}


\maketitle

\section*{Overview}
This supplementary material introduces experiment details, details of \textit{prompt generator}, additional analysis, test set, and further qualitative results.
\begin{itemize}
\item Section \hyperref[sup:exp]{A} provides more experiment details about the baseline, metrics, and human evaluation.
\item Section \hyperref[sup:generator]{B} presents the usage of \textit{prompt generator}, including full descriptions used to generate individual prompts from the scenario containing the sequence of events.
\item Section \hyperref[sup:cost]{C} provides an additional analysis of computational cost and hyper-parameters.
\item Section \hyperref[sup:experiment]{D} provides details of evaluation including test set.
\item Section \hyperref[sup:qualitative]{E} provides more qualitative results in diverse domains. We provide more comparison results between state-of-the-art models and qualitative results. Furthermore, we also present more generated videos conditioning on image and multi-text and examples generated by the prompt generator.
\end{itemize}
\section*{A. Experiment Details}
\label{sup:exp}
\myparagraph{Baselines.}
To demonstrate the effectiveness of our proposed MEVG, we compare the outcomes with several existing baselines. We select baselines that enable synthesizing the videos with multiple prompts without any training or fine-tuning.
DirecT2V~\cite{hong2023large} and Text2Video-Zero~(T2V-Zero)~\cite{khachatryan2023text2video} leverage Stable Diffusion~\cite{rombach2022high} trained on only text-image pairs.
These models utilize the frame-level descriptions to create individual frames constituting the video content.
Furthermore, two text-to-video-based methods, Gen-L-Video~\cite{wang2023gen} and VidRD~\cite{gu2023reuse}, are used to compare with ours. 
We use LVDM ~\cite{he2022latent} as the foundation model for our experiment to be a fair comparison.

\myparagraph{Metrics.}
We report CLIP Similarity (``\textbf{CLIP-Text}") \cite{radford2021learning, hessel2021clipscore} and temporal consistency (``\textbf{CLIP Image}") \cite{esser2023structure, qi2023fatezero, Ceylan_2023_ICCV, wu2023tune} to evaluate our proposed MEVG. 
CLIP-Text is a commonly employed metric to measure the correlation between two different modalities, image and text.
We compute the cosine similarity over all frames corresponding to each prompt to present how well the outcomes reflect the meaning of given conditions. Additionally, CLIP-Image is used to measure the correlation of frame. We compute the cosine similarity between two consecutive frames and take the average value over all frames.
Furthermore, we conduct a human evaluation to measure four properties of outcomes: temporal consistency, semantic alignment, realism, and preference. 

\myparagraph{Human Evaluation.}
We conduct a human evaluation study to measure four properties of outcomes: temporal consistency, semantic alignment, realism, and preference. 
Specifically, we request all participants assign a score on a scale 1~(low quality) to 5~(high quality) for the following set of four questions. 
First,``\textit{How smoothly the content of videos changes in response to the given prompts.}" indicates how each video clip is smoothly connected between the distinct prompts (Temporal Consistency). Second,``\textit{How well does the video correspond with the prompts.}" evaluates how well the generated video reflects a given sequence of prompts~(Semantic Alignment). Third, ``\textit{How natural and real does this video look, considering the consistency of the background and the objects.}" evaluates the realism of the generated video concerning the background and object consistency~(Realism). Finally, ``\textit{Considering the three questions above, please rank the overall video quality.}" leads to participants ranking their preference over the generated video based on comprehensive perspective~(Preference).

\begin{figure*}[b!]
  \centering
  \includegraphics[width=\textwidth]{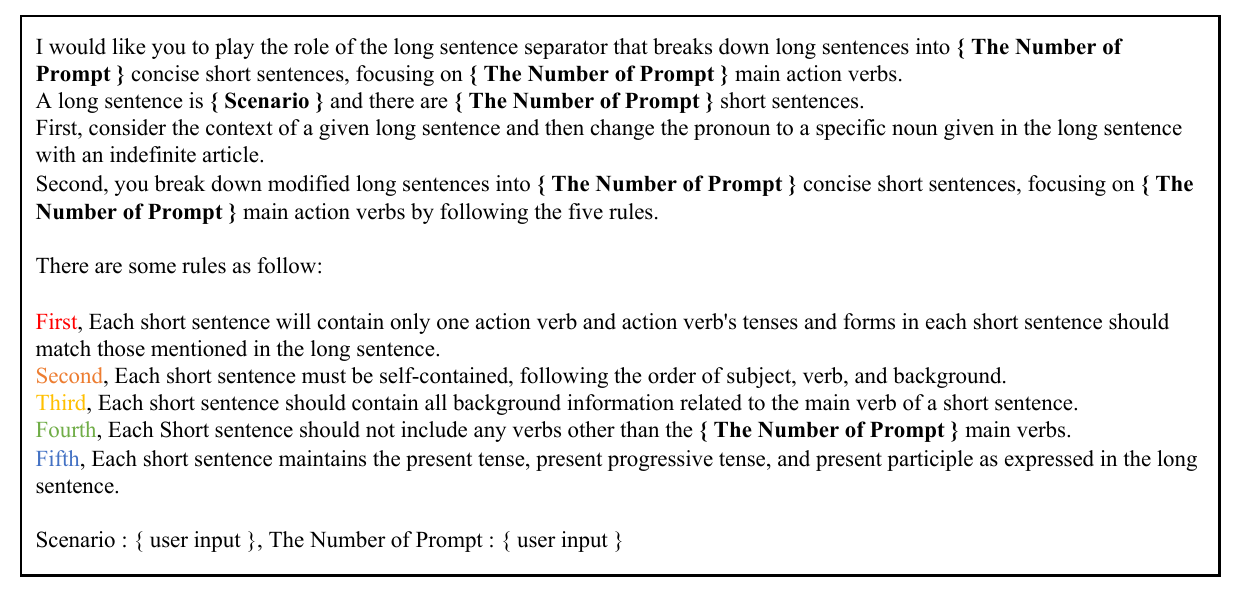}
  \caption{This instruction follows the five guidelines to create individual prompts based on a given scenario and the number of prompts by the user.}
  \label{fig:instruction}
\end{figure*}

\section*{B. Details of Prompt Generator}
\label{sup:generator}
In this section, we provide additional information of \textit{prompt generator} that is described in our main paper~(see Sec. 3.5). 
Our \textit{prompt generator} can naturally split a single scenario containing multiple events into distinct prompts with a prescribed textural format. 

\myparagraph{Instruction for \textit{prompt generator}}
The key point of the \textit{prompt generator} is that each prompt has a single event while maintaining the comprehensive content of the scenario. Inspired by the Free-Bloom~\cite{huang2023free} and DirecT2V~\cite{hong2023large}, we devise adequate instruction following the five concrete rules~(see Fig.~\ref{fig:instruction}).

\section*{C. Additional Analysis}
\label{sup:cost}

\noindent
\vspace{-2em}
\begin{wrapfigure}{hr}{0.4\textwidth}
\includegraphics[width=\linewidth]{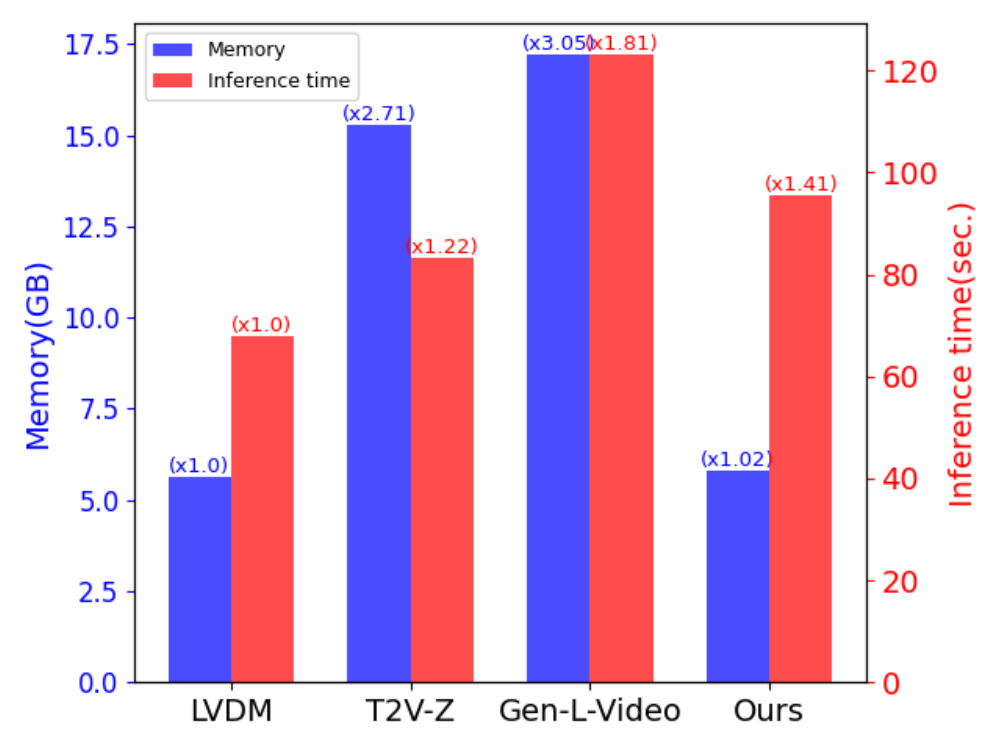}
    \caption{Analysis on additional cost}
    \label{fig:cost}
\end{wrapfigure}

\myparagraph{Analysis on Additional Cost.} We have analyzed additional computation costs over the three prompts on a single NVIDIA GeForce RTX 3090.
As shown in Fig.~\ref{fig:cost}, we figure out that our method exhibits a marginal increase in memory usage~($\times$1.02) and inference time~($\times$1.41) in comparison to the base model, LVDM~\cite{he2022latent}. 
However, Gen-L-Video~\cite{wang2023gen}, which uses the same base model, requires significantly greater resources~($\times$3.05 / $\times$1.81). 
Furthermore, despite T2V-Zero~\cite{khachatryan2023text2video} generating only 8 frames per prompt instead of 16 frames, our approach demonstrates comparable speed.

\myparagraph{Hyper-parameter Analysis.} We present the analysis on the hyper-parameters $\delta_{\text{LFAI}}$ and $\delta_{\text{SGS}}$ as shown in Tab.~\ref{sup:tab1}. 
We report the automatic metrics~(e.g. CLIP-Text and CLIP-Image) according to the variation of hyper-parameters, measured by five samples per scenario. 
Essentially, we observe that high values of $\delta_{\text{LFAI}}$ and $\delta_{\text{SGS}}$ demonstrate strong visual coherence but a decline in semantic alignment.

\begin{table*}[h!]
\caption{Effect of two independent hyper-parameters $\delta_{\text{LFAI}}$ and $\delta_{\text{SGS}}$.}
\label{sup:tab1}
    \begin{minipage}{.5\textwidth}\
      \centering
        \begin{tabular}{ccc}
        \toprule
        $\delta_{\text{LFAI}}$ & CLIP-Text & CLIP-Image \\
        \midrule
        1 & 30.399 & 0.938 \\
        10 & 30.397 & 0.939 \\
        100 & 30.275 & 0.941 \\
        500 & 30.600 & 0.944 \\
        1000 & 30.674 & 0.945 \\
        \bottomrule
      \end{tabular}
    \end{minipage}%
    \begin{minipage}{.5\textwidth}
      \centering
        \begin{tabular}{ccc}
        \toprule
        $\delta_{\text{SGS}}$ & CLIP-Text & CLIP-Image \\
        \midrule
        0.01 & 30.836 & 0.933 \\
        0.1 & 30.870 & 0.935 \\
        7 & 30.674 & 0.945 \\
        15 & 30.530 & 0.941 \\
        50 & 30.102 & 0.954 \\
        \bottomrule
      \end{tabular}
    \end{minipage} 
\end{table*}

\section*{D. Details of Evaluation}
\label{sup:experiment}

\myparagraph{Test Set}
\label{sec:testset}
Since there are no evaluation datasets for multi-text-based video generation reflecting multiple events, we construct a test set by referring to generative model literature communities.
Some prompts are derived from the existing works~\cite{huang2023free, khachatryan2023text2video, qiu2023freenoise}.
To evaluate the quality of the generated videos, we design complex scenarios consisting of multiple prompts. Each scenario is divided into three categories: background transitions, object movements,  and complex content changes.
Scenarios consist of two, three, or four prompts while containing diverse objects and backgrounds in different domains. The test set is listed as follows:

\myparagraph{Background Transition}

\begin{itemize}
    \item \textit{Scenario} 1.
    
    \noindent\line(1,0){230}
    \begin{enumerate}
        \item The teddy bear goes under water in San Francisco.
        \item The teddy bear keeps swimming under the water with colorful fishes.
        \item A teddy bear is swimming under water.
        
    \end{enumerate}
    \vspace{1em}
    \item \textit{Scenario} 2.
    
    \line(1,0){230}
    \begin{enumerate}
        \item An astronaut in a white uniform is snowboarding in the snowy hill.
        \item An astronaut in a white uniform is surfing in the sea.
        \item An astronaut in a white uniform is surfing in the desert.
    \end{enumerate}
    \vspace{1em}
    \item \textit{Scenario} 3.
    
    \line(1,0){230}
    \begin{enumerate}
        \item A white butterfly sits on a purple flower.
        \item The color of the purple flower where the white butterfly sits turns red.
        \item A white butterfly is sitting on a red flower.
    \end{enumerate}
    \vspace{1em}
    \item \textit{Scenario} 4.
    
    \line(1,0){230}
    \begin{enumerate}
        \item The caterpillar is on the leaves.
        \item The caterpillar eats the leaves.
        \item The caterpillar ate all the leaves.
    \end{enumerate}
    \vspace{1em}
    \item \textit{Scenario} 5.
    
    \line(1,0){230}
    \begin{enumerate}
    \item The teddy bear is swimming under the sea.
    \item The teddy bear is playing with colorful fishes while swimming under the sea.
    \item The teddy bear is resting quietly among the coral reefs under the sea.
    \item Suddenly a shark appeared next to the teddy bear under the sea.
    \end{enumerate}
    \vspace{1em}
    \item \textit{Scenario} 6.
    
    \line(1,0){230}
    \begin{enumerate}
        \item A man runs the starry night road in Van Gogh style.
        \item A man runs the starry night road in Monet style.
        \item A man runs the starry night road in Picasso style.
        \item A man runs the starry night road in Da Vinci style.
    \end{enumerate}
    \vspace{1em}
    \item \textit{Scenario} 7.
    
    \line(1,0){230}
    \begin{enumerate}
        \item The whole beautiful night view of the city is shown.
        \item Heavy rain flood the city with beautiful night scenery and flood.
        \item The day dawns over the flooded city.
    \end{enumerate}
    \vspace{1em}
    \item \textit{Scenario} 8.
    
    \line(1,0){230}
    \begin{enumerate}
        \item Cherry blossoms bloom around the Japanese-style castle.
        \item Leaves fall around the Japanese-style castle.
        \item Snow falls around the Japanese-style castle.
        \item Snow builds up in trees around the Japanese-style castle.
    \end{enumerate}
    \vspace{1em}
    \item \textit{Scenario} 9.
    
    \line(1,0){230}
    \begin{enumerate}
        \item The dog is standing on Times Square Street.
        \item The dog is standing on the Japanese street.
        \item The dog is standing on the China town.
        \item The dog is standing on the street in Korea.
    \end{enumerate}
    \vspace{1em}
    \item \textit{Scenario} 10.
    
    \line(1,0){230}
    \begin{enumerate}
        \item In spring, a white butterfly sit on a flower.
        \item In summer, a white butterfly sit on flower.
        \item In autumn, a white butterfly sit on flower.
        \item In winter, a white butterfly sit on flower.
    \end{enumerate}

\end{itemize}

\myparagraph{Object Motion}
\vspace{0.5em}
\begin{itemize}
    \item \textit{Scenario} 11.
    
    \line(1,0){230}
    \begin{enumerate}
        \item Two men play tennis in the green gym.
        \item Two men playing tennis swing a racket in the green gym.
        \item A tennis ball passes between two men playing tennis in the green gym.
    \end{enumerate}
    \vspace{1em}
    \item \textit{Scenario} 12.
    
    \line(1,0){230}
    \begin{enumerate}
        \item A man sits in front of a standing microphone on Times Square Street and plays the guitar.
        \item The man sits on the street in Times Square and sings on the guitar.
        \item The man sits on Times Square Street and keeps playing the guitar.
    \end{enumerate}
    \vspace{1em}
    \item \textit{Scenario} 13.
    
    \line(1,0){230}
    \begin{enumerate}
        \item A shark swims with colorful fish in the sea.
        \item A shark swims with scuba divers in the sea.
        \item A shark dances with scuba divers in the sea.
    \end{enumerate}
    \vspace{1em}
    \item \textit{Scenario} 14.
    
    \line(1,0){230}
    \begin{enumerate}
        \item A candle is brightly lit in the dark room.
        \item Smoke rises from an unlit candle in the dark room.
        \item There is an unlit candle in a dark room.
    \end{enumerate}
    \vspace{1em}
    \item \textit{Scenario} 15.
    
    \line(1,0){230}
    \begin{enumerate}
        \item There is a beach where there is no one.
        \item The waves hit the deserted beach.
        \item There is a beach that has been swept away by waves.
    \end{enumerate}
    \vspace{1em}
    \item \textit{Scenario} 16.
    
    \noindent\line(1,0){230}
    \begin{enumerate}
        \item A dog runs in the snowy mountains.
        \item A dog barks on snowy mountain.
        \item A dog stands on snowy mountain.
        \item A dog lies down on the snowy mountain.
    \end{enumerate}
    \vspace{1em}
    \item \textit{Scenario} 17.
    
    \line(1,0){230}
    \begin{enumerate}
        \item A man runs on a beautiful tropical beach at sunset of 4k high resolution.
        \item A man rides a bicycle on a beautiful tropical beach at sunset of 4k high resolution.
        \item A man walks on a beautiful tropical beach at sunset of 4k high resolution.
        \item A man reads a book on a beautiful tropical beach at sunset of 4k high resolution.
    \end{enumerate}
    \vspace{1em}
    \item \textit{Scenario} 18.
     
    \line(1,0){230}
    \begin{enumerate}
        \item A sheep is standing in a field full of grass.
        \item A sheep graze in a field full of grass.
        \item A sheep is running in a field full of grass.
        \item A sheep is lying in a field full of grass.
    \end{enumerate}
    \vspace{1em}
    \item \textit{Scenario} 19.
    
    \line(1,0){230}
    \begin{enumerate}
        \item A golden retriever has a picnic on a beautiful tropical beach at sunset.
        \item A golden retriever is running towards a beautiful tropical beach at sunset.
        \item A golden retriever sits next to a bonfire on a beautiful tropical beach at sunset.
        \item A golden retriever is looking at the starry sky on a beautiful tropical beach.
    \end{enumerate}
    \vspace{1em}
    \item \textit{Scenario} 20.
    
    \line(1,0){230}
    \begin{enumerate}
        \item A Red Riding Hood girl walks in the woods.
        \item A Red Riding Hood girl sells matches in the forest.
        \item A Red Riding Hood girl falls asleep in the forest.
        \item A Red Riding Hood girl walks towards the lake from the forest.
    \end{enumerate}
    
\end{itemize}

\myparagraph{Complex content changes}
\vspace{0.5em}
\begin{itemize}
    \item \textit{Scenario} 21.
    
    \noindent\line(1,0){230}
    \begin{enumerate}
        \item Side view of an astronaut is walking through a puddle on mars.
        \item The astronaut watches fireworks.
    \end{enumerate}
    \vspace{1em}
    \item \textit{Scenario} 22.
    
    \line(1,0){230}
    \begin{enumerate}
        \item The astronaut gets on the spacecraft.
        \item The spacecraft goes from Earth to Mars.
        \item The spacecraft lands on Mars.
    \end{enumerate}
    \vspace{1em}
    \item \textit{Scenario} 23.
    
    \line(1,0){230}
    \begin{enumerate}
        \item The volcano erupts in the clear weather.
        \item Smoke comes from the crater of the volcano, which has ended its eruption in the clear weather.
        \item The weather around the volcano turns cloudy.
    \end{enumerate}
    \vspace{1em}
    \item \textit{Scenario} 24.
    
    \line(1,0){230}
    \begin{enumerate}
        \item There is a Mickey Mouse dancing through the spring forest.
        \item There is a Mickey Mouse walking through the autumn forest.
        \item There is a Mickey Mouse running through the winter forest.
    \end{enumerate}
    \vspace{1em}
    \item \textit{Scenario} 25.
    
    \line(1,0){230}
    \begin{enumerate}
        \item A panda is playing guitar on Times Square.
        \item The panda is singing on Times Square.
        \item The panda starts dancing.
        \item People in Times Square clap for the panda.
    \end{enumerate}
    \vspace{1em}
    \item \textit{Scenario} 26.
    
    \line(1,0){230}
    \begin{enumerate}
        \item A teddy bear walks on the streets of Times Square .
        \item The teddy bear enters restaurants.
        \item The teddy bear eats pizza.
        \item The teddy bear drinks water.
    \end{enumerate}
    \vspace{1em}
    \item \textit{Scenario} 27.
     
    \line(1,0){230}
    \begin{enumerate}
        \item The cartoon-style bear appears in a comic book.
        \item The cartoon-style bears in comic books jump out into the real world.
        \item The bear in the real world dances.
        \item The bear in the real world sits.
    \end{enumerate}
    \vspace{1em}
    \item \textit{Scenario} 28.
    
    \line(1,0){230}
    \begin{enumerate}
        \item A chihuahua in astronaut suit floating in space, cinematic lighting, glow effect.
        \item A chihuahua in astronaut suit dancing in space, cinematic lighting, glow effect.
        \item A chihuahua in astronaut suit swimming under the water, clean, brilliant effect.
        \item A chihuahua in astronaut suit swimming under the water with colorful fishes, clean, brilliant effect.
    \end{enumerate}
    \vspace{1em}
    \item \textit{Scenario} 29.
    
    \line(1,0){230}
    \begin{enumerate}
        \item A waterfall flows in the mountains under a clear sky.
        \item A waterfall flows in the fall mountains under a clear sky.
        \item A waterfall flows in the winter mountains under a clear sky.
        \item A waterfall frozen on a mountain during a snowstorm.
    \end{enumerate}
    \vspace{1em}
    \item \textit{Scenario} 30.
    
    \line(1,0){230}
    \begin{enumerate}
        \item The boulevards are quiet in the clear sky.
        \item The boulevards are quiet in the night sky.
        \item The boulevards are crowded in the night sky.
        \item The boulevards are crowded under the firework sky. 
    \end{enumerate}
    
\end{itemize}

\section*{E. Qualitative Results}
\label{sup:qualitative}
\begin{figure*}[t]
\includegraphics[width=\textwidth]{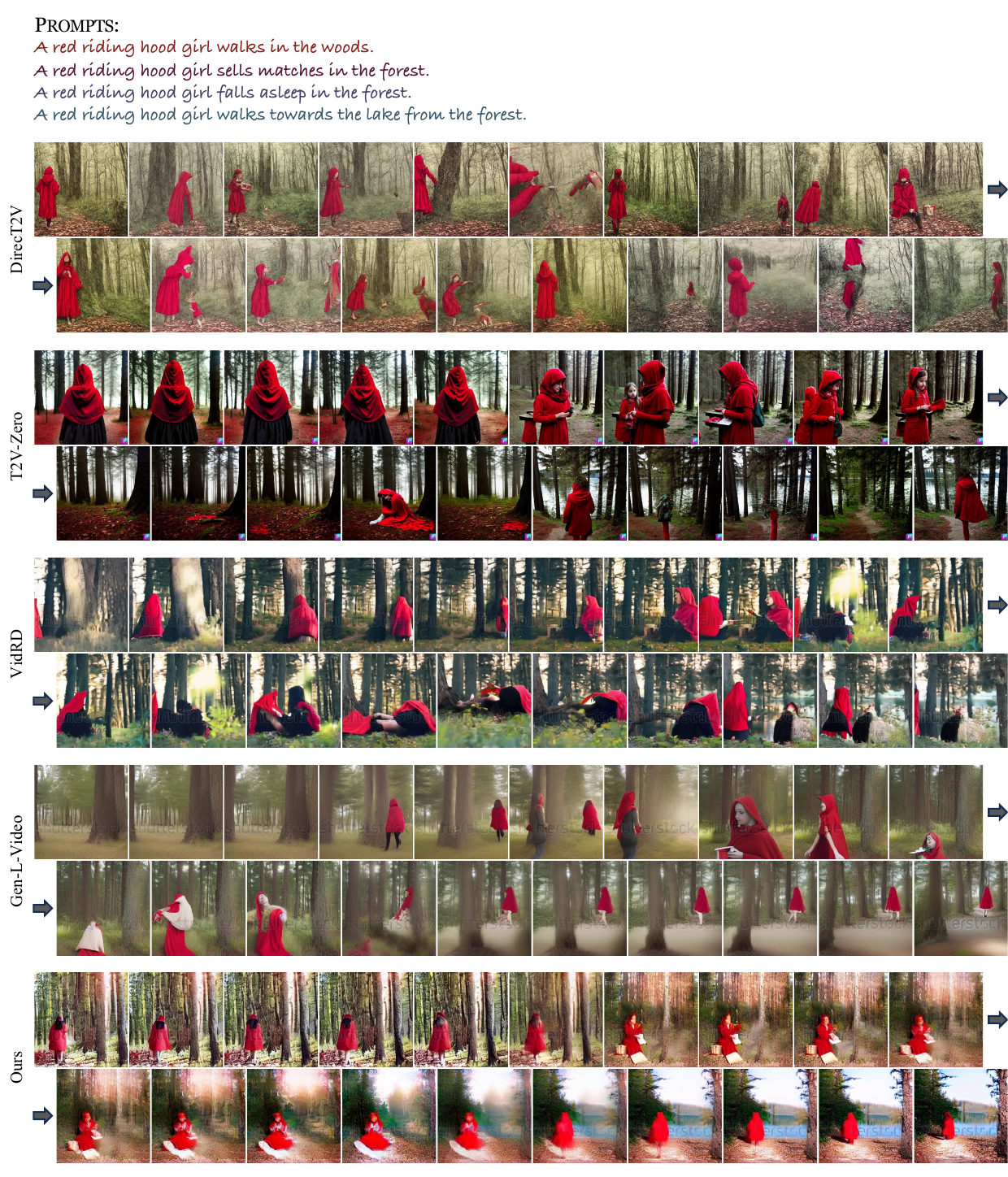}
\captionof{figure}
{Qualitative comparisons with DirecT2V~\cite{hong2023large}, T2V-Zero~\cite{khachatryan2023text2video}, VidRD~\cite{gu2023reuse}, and Gen-L-Video~\cite{wang2023gen}}
\label{fig:sup_comparison_red}
\end{figure*}

\begin{figure*}[t]
\includegraphics[width=\textwidth]{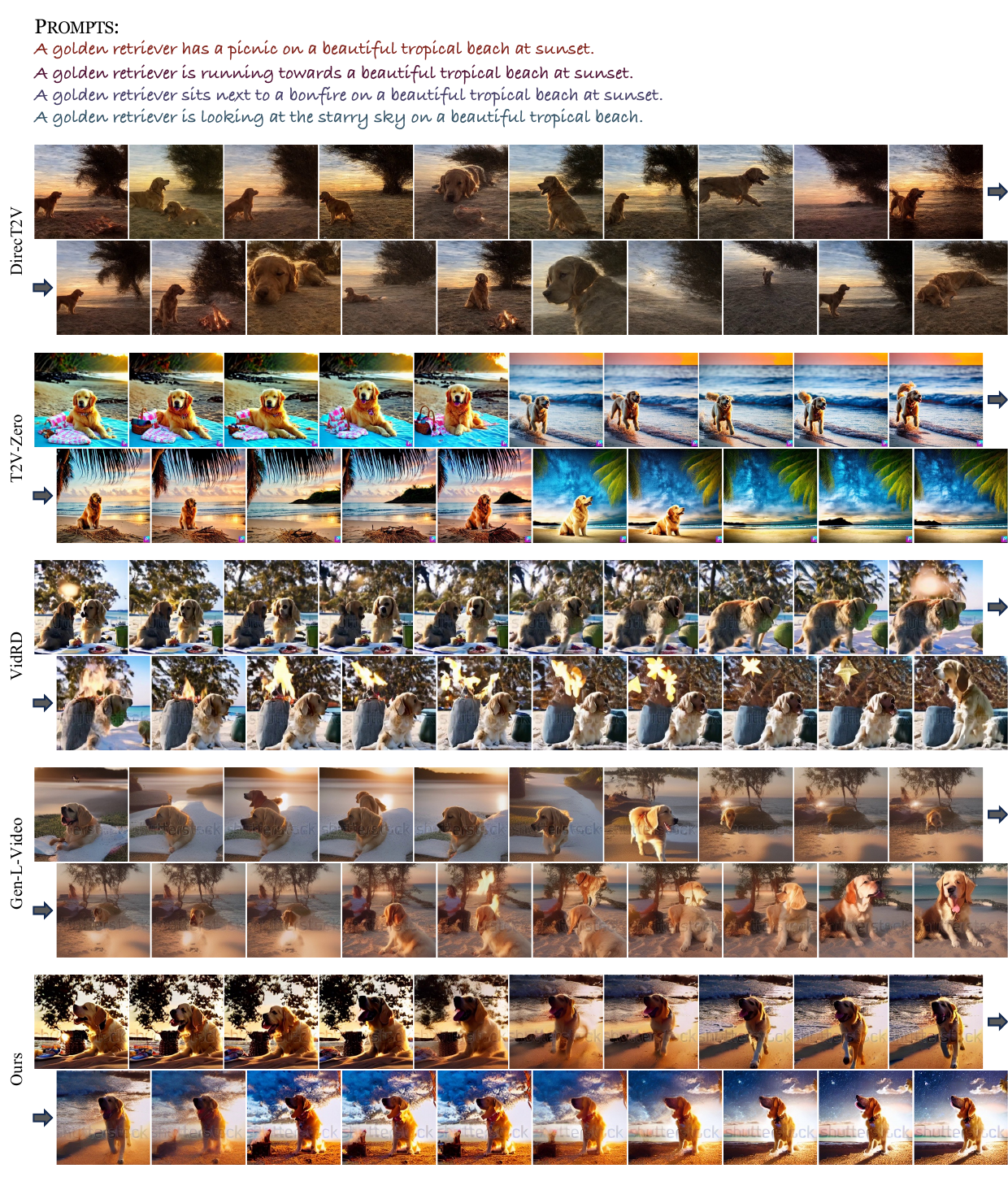}
\captionof{figure}
{Qualitative comparisons with DirecT2V~\cite{hong2023large}, T2V-Zero~\cite{khachatryan2023text2video}, VidRD~\cite{gu2023reuse}, and Gen-L-Video~\cite{wang2023gen}}
\label{fig:sup_comparison_4kdog}
\end{figure*}

\begin{figure*}[t]
\includegraphics[width=\textwidth]{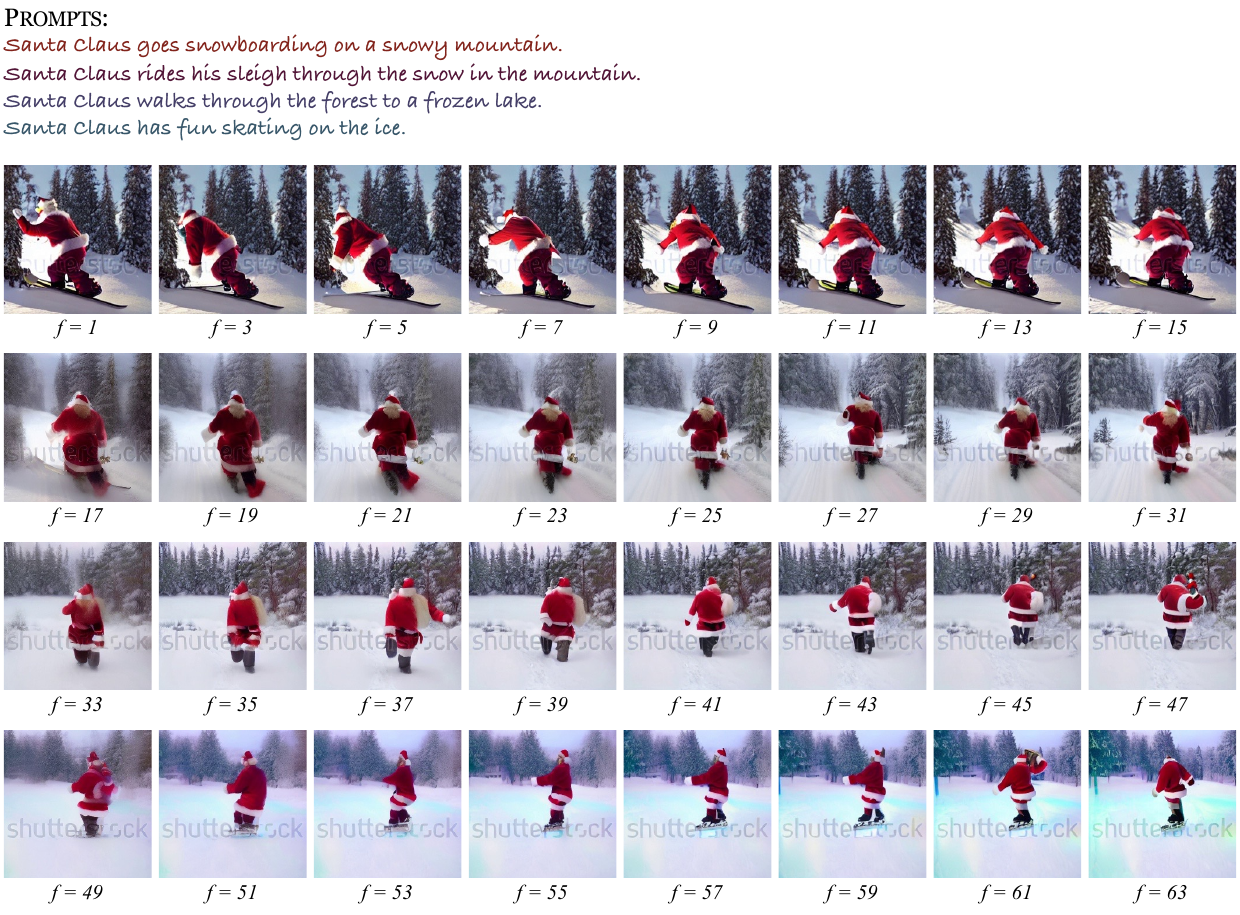}
\captionof{figure}
{Qualitative result conditioning on multi-text with LVDM~\cite{he2022latent}.}
\label{fig:sup_qualitative_santa}
\end{figure*}

\begin{figure*}[t]
\includegraphics[width=\textwidth]{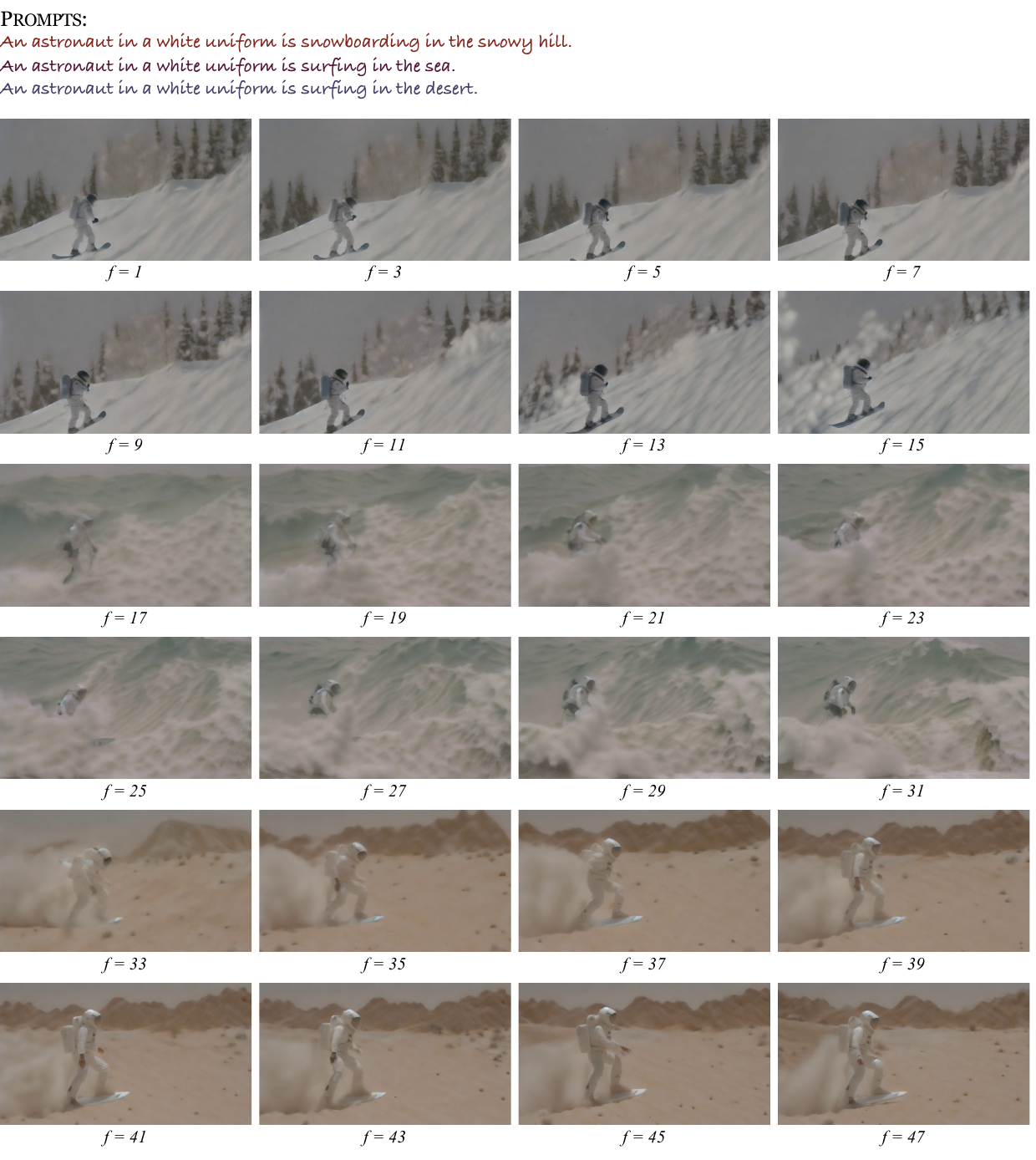}
\captionof{figure}
{Qualitative result conditioning on multi-text with VideoCrafter1~\cite{chen2023videocrafter1}.}
\label{fig:sup_videocrafter_ski}
\end{figure*}

\begin{figure*}[t]
\begin{center}
\includegraphics[width=0.95\textwidth,height=0.95\textheight]{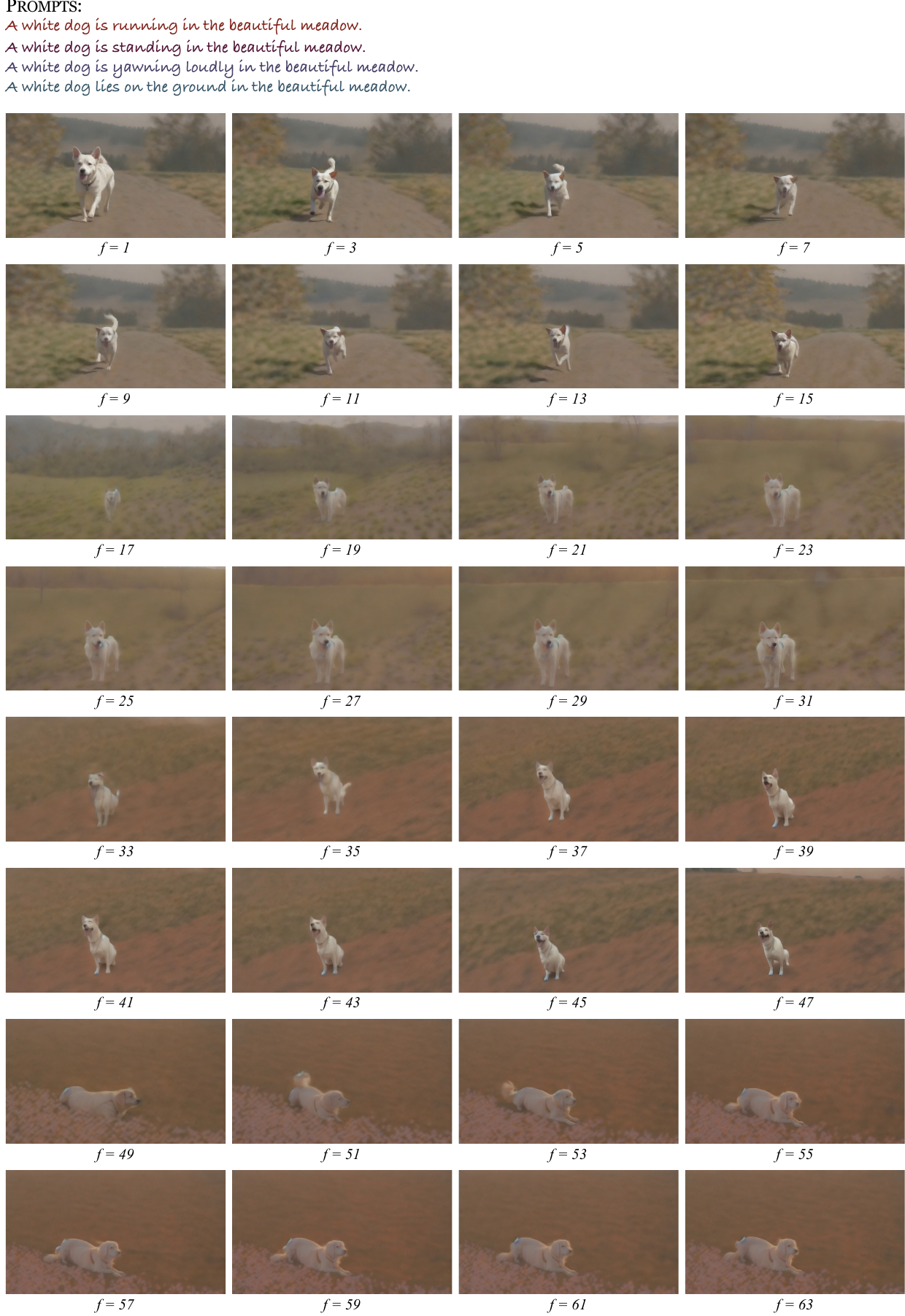}
\captionof{figure}
{Qualitative result conditioning on multi-text with VideoCrafter1~\cite{chen2023videocrafter1}.}
\label{fig:sup_videocrafter_dog}   
\end{center}
\end{figure*}

\begin{figure*}[t]
\includegraphics[width=\textwidth]{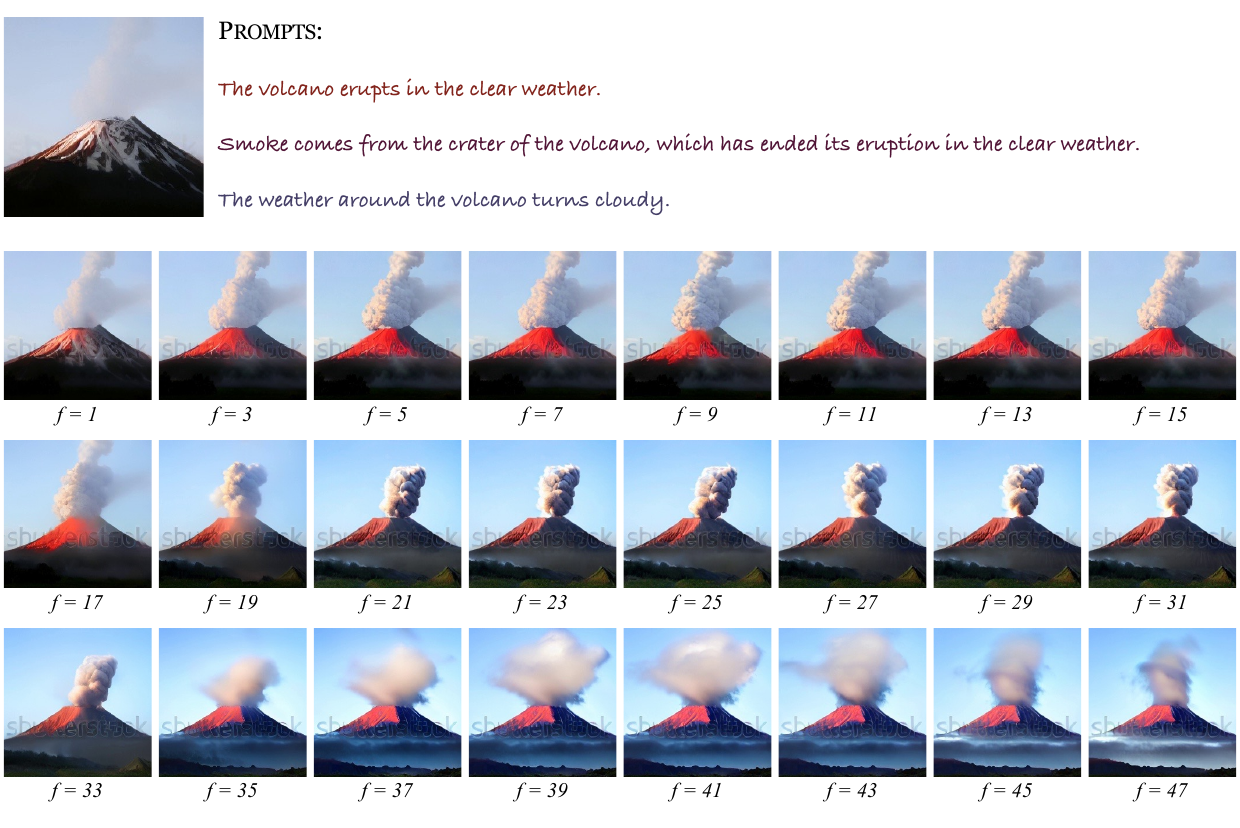}
\captionof{figure}
{Example of generated video conditioning on image and multi-text.}
\label{fig:sup_image_fuzi}
\end{figure*}

\begin{figure*}[t]
\includegraphics[width=\textwidth]{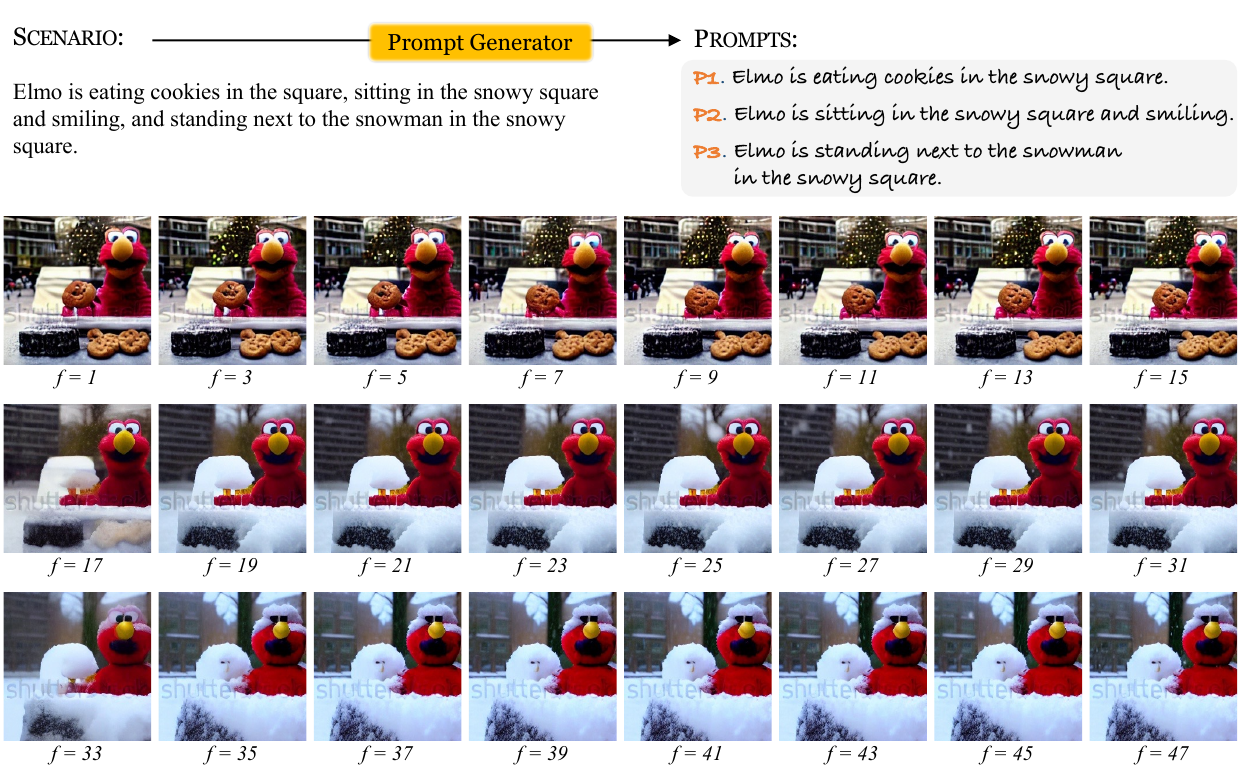}
\captionof{figure}
{Examples of multi-text video generation utilizing the \textit{prompt generator}.}
\label{fig:sup_promptgenerator_elmo}
\end{figure*}

\begin{figure*}[t]
\includegraphics[width=\textwidth]{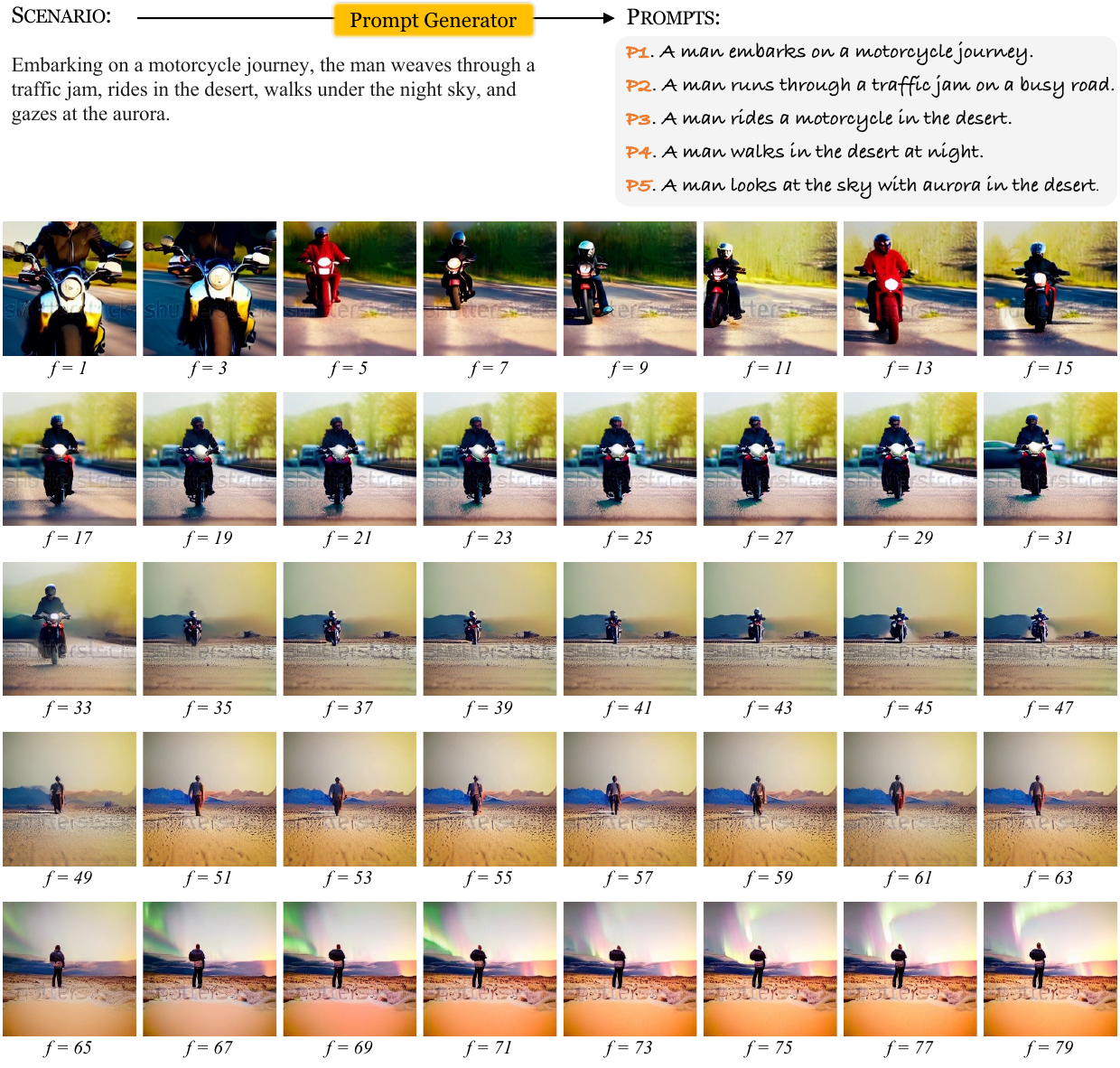}
\captionof{figure}
{Examples of multi-text video generation utilizing the \textit{prompt generator}.}
\label{fig:sup_promptgenerator_motor}
\end{figure*}

In this section, we provide more qualitative results of our methods in the multi-text video generation setting. Specifically, Fig.~\ref{fig:sup_comparison_red} and Fig.~\ref{fig:sup_comparison_4kdog} represent the qualitative comparison with the state-of-the-art methods~\cite{hong2023large, khachatryan2023text2video, gu2023reuse, wang2023gen}. In Fig.~\ref{fig:sup_qualitative_santa} $\sim$ Fig.~\ref{fig:sup_videocrafter_dog}, we showcase the multi-text video generation results over the diverse domain. Nota that, Fig~\ref{fig:sup_videocrafter_ski} and Fig~\ref{fig:sup_videocrafter_dog} leverage the different foundation model\footnote{VideoCrafter1 ~\cite{chen2023videocrafter1} is used as the foundation model in this experiment.} and generate 16 frames per each prompt with 576$\times$1024 resolution.
Furthermore, we visualize the generated videos conditioning on image and multi-text~(see Fig.~\ref{fig:sup_image_fuzi}). Finally, we present additional results generated by the \textit{prompt generator} in Fig.~\ref{fig:sup_promptgenerator_elmo} and Fig.~\ref{fig:sup_promptgenerator_motor}.

\begin{figure*}[t]
\includegraphics[width=\textwidth]{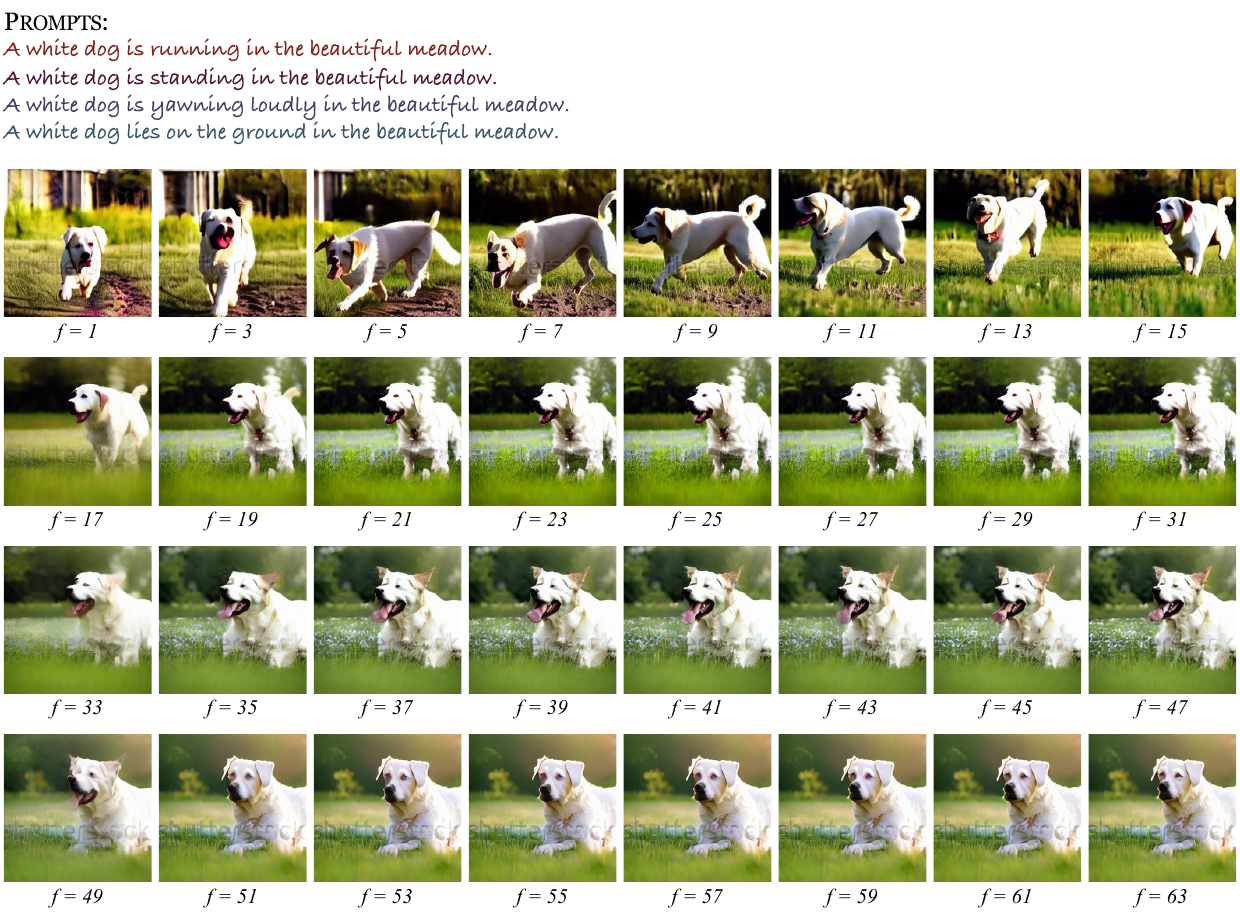}
\captionof{figure}
{Qualitative result conditioning on multi-text with LVDM~\cite{he2022latent}.}
\label{fig:sup_qualitative_dog}
\end{figure*}

\begin{figure*}[t]
\includegraphics[width=\textwidth]{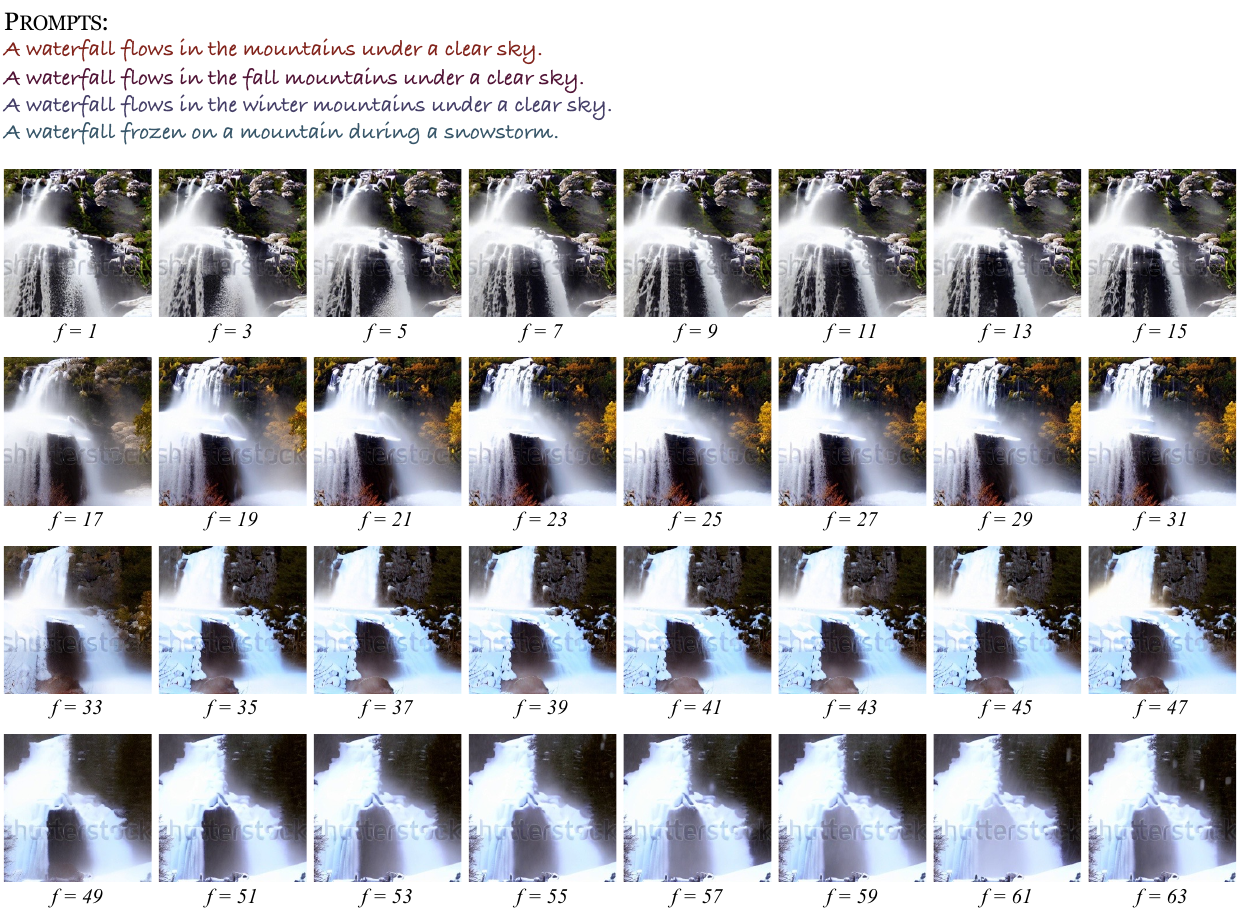}
\captionof{figure}
{Qualitative result conditioning on multi-text with LVDM~\cite{he2022latent}.}
\label{fig:sup_qualitative_waterfall}
\end{figure*}

\clearpage

%
%

\end{document}